\documentclass[preprint,3p]{elsarticle}
\usepackage{amsmath,amsfonts,amssymb}
\usepackage{lineno,enumitem}
\usepackage[usenames,dvipsnames,svgnames,table]{xcolor}
\usepackage{color,soul}
\usepackage{tcolorbox}
\usepackage{amssymb}
\usepackage{pifont}
\usepackage{makecell}
\usepackage{booktabs}
\usepackage{amsmath,amssymb}
\usepackage[ruled,vlined]{algorithm2e}
\usepackage{graphicx}
\usepackage{url}
\usepackage{lineno}
\usepackage{color}
\usepackage{subfig}
\usepackage{amsmath}
\usepackage{enumitem}
\usepackage{multirow}
\usepackage{subfig}
\usepackage{tikz,forest}
\usepackage{booktabs}
\usepackage{pdflscape}
\usepackage{array}
\usepackage{makecell}

\newtheorem{definition}{Definition}

\newcolumntype{L}[1]{>{\raggedright\let\newline\\\arraybackslash\hspace{0pt}}m{#1}}
\newcolumntype{C}[1]{>{\centering\let\newline\\\arraybackslash\hspace{0pt}}m{#1}}
\newcolumntype{R}[1]{>{\raggedleft\let\newline\\\arraybackslash\hspace{0pt}}m{#1}}

\DeclareMathOperator*{\argmin}{arg\,min}

\xdefinecolor{gray95}{gray}{0.65}
\xdefinecolor{gray25}{gray}{0.1}

\usepackage{scalerel}

\modulolinenumbers[5]

\usepackage{tikz}

\journal{undisclosed journal}

\begin{document}
	
\begin{frontmatter}
	
	\title{On the Post-hoc Explainability of Deep Echo State Networks for Time Series Forecasting, Image and Video Classification}
	
	\author[tec]{Alejandro Barredo Arrieta}
	\author[tec]{Sergio Gil-Lopez}
	\author[tec]{Ibai La\~na}
	\author[upv]{Miren Nekane Bilbao}
	\author[tec,upv]{Javier~Del Ser\corref{cor1}}
	
	\cortext[cor1]{Corresponding author. TECNALIA, Basque Research and Technology Alliance (BRTA), 48160 Derio, Spain. Phone: +34 664 113 013. Email address: javier.delser@tecnalia.com}
	\address[tec]{TECNALIA, Basque Research and Technology Alliance (BRTA), 48160 Derio, Spain}
	\address[upv]{University of the Basque Country (UPV/EHU), 48013 Bilbao, Spain}
	
	\begin{abstract}
		Since their inception, learning techniques under the Reservoir Computing paradigm have shown a great modeling capability for recurrent systems without the computing overheads required for other approaches, specially deep neural networks. Among them, different flavors of echo state networks have attracted many stares through time, mainly due to the simplicity and computational efficiency of their learning algorithm. However, these advantages do not compensate for the fact that echo state networks remain as black-box models whose decisions cannot be easily explained to the general audience. This issue is even more involved for multi-layered (also referred to as \emph{deep}) echo state networks, whose more complex hierarchical structure hinders even further the explainability of their internals to users without expertise in Machine Learning or even Computer Science. This lack of explainability can jeopardize the widespread adoption of these models in certain domains where accountability and understandability of machine learning models is a must (e.g. medical diagnosis, social politics). This work addresses this issue by conducting an explainability study of Echo State Networks when applied to learning tasks with time series, image and video data. Specifically, the study proposes three different techniques capable of eliciting understandable information about the knowledge grasped by these recurrent models, namely, potential memory, temporal patterns and pixel absence effect. Potential memory addresses questions related to the effect of the reservoir size in the capability of the model to store temporal information, whereas temporal patterns unveils the recurrent relationships captured by the model over time. Finally, pixel absence effect attempts at evaluating the effect of the absence of a given pixel when the echo state network model is used for image and video classification. We showcase the benefits of our proposed suite of techniques over three different domains of applicability: time series modeling, image and, for the first time in the related literature, video classification. Our results reveal that the proposed techniques not only allow for a informed understanding of the way these models work, but also serve as diagnostic tools capable of detecting issues inherited from data (e.g. presence of hidden bias).
	\end{abstract}
	
	
\end{frontmatter}

\section{Introduction} \label{sec:intro}

Since their inception \cite{jaeger2003adaptive,lukovsevivcius2009reservoir}, Echo State Networks (ESNs) have been frequently proposed as an efficient replacement for traditional Recurrent Neural Networks (RNNs). As opposed to conventional gradient-based RNN training, the recurrent part (reservoir) of ESNs is not updated via gradient backpropagation, but is simply initialized at random, given that certain mathematical properties are met. As a result, their training overhead for real life applications becomes much less computationally demanding than that of RNNs. Over the well-known Mackey-Glass chaotic time series prediction benchmark, ESN has been shown to improve the accuracy scores achieved by multi-layer perceptrons (MLPs), support vector machines (SVMs), backpropagation-based RNNs and other learning approaches by a factor of over 2000 \cite{jaeger2004harnessing}. These proven benefits have appointed ESN as a top contending dynamical model for performance and computational efficiency reasons when compared to other modeling counterparts.

Unfortunately, choosing the right parameters to initialize this reservoir falls a bit on the side of luck and past experience of the scientist \cite{wu2018statistical}, and less on that of sound reasoning. As stated in \cite{jaeger2005reservoir} and often referred thereafter, the current approach for assessing whether a reservoir is suited for a particular task is to observe if it yields accurate results, either by handcrafting the values of the reservoir parameters or by automating their configuration via an external optimizer. All in all, this poses tough questions to address when developing an ESN for a certain application, since knowing whether the created structure is optimal for the problem at hand is not possible without actually training it. Furthermore, despite recent attempts made in this direction \cite{thiede2019gradient,ozturk2020optimizing}, there is no clear consensus on how to guide the search for good reservoir based models.

Concerns in this matter go a step beyond the ones exposed above about the configuration of these models. Model design and development should orbit around a deep understanding of the multiple factors hidden below the surface of the model. For this purpose, a manifold of techniques have been proposed under the Explainable Artificial Intelligence (XAI) paradigm for easing the understanding of decisions issued by existing AI-based models. The information delivered by XAI techniques allow improving the design/configuration of AI models, extracting augmented knowledge about their outputs, accelerating debugging processes, or achieving a better outreach and adoption of this technology by non-specialized audience \cite{arrieta2020explainable}. Although the activity in this research area has been vibrant for explaining many black-box machine learning models, to the best of our knowledge there is no prior work on the development of techniques of this sort for dynamical approaches. The need for providing explanatory information about the knowledge learned by ESNs remains unaddressed, even though recent advances on the construction of multi-layered reservoirs (Deep ESN \cite{gallicchio2017deep}) that make these models more opaque than their single-layered counterparts.

Given the above context, this manuscript takes a step ahead by presenting a novel suite of XAI techniques suited to issue explanations of already trained Deep ESN models. The proposed techniques elicit visual information that permit to assess the memory properties of these models, visualize their detected patterns over time, and analyze the importance of a individual input in the model's output. We provide mathematical definitions of the XAI factors involved in the tools of the proposed framework, complemented by examples that help illustrate their applicability and comprehensibility to the general audience. This introduction of the overall framework is complemented by several experiments showcasing the use of our XAI framework to three different real applications: 1) battery cell consumption prediction, 2) road traffic flow forecasting, 3) image classification and 4) video classification. The results are conclusive: the outputs of the proposed XAI techniques confirm, in a human-readable fashion, that the Deep ESN models capture temporal dependencies existing in data that could be expected due to prior knowledge, and that this captured information can be summarized to deepen the understanding of a general practitioner/user consuming the model's output. 

The rest of the paper is organized as follows: Section \ref{sec:background} provides the reader with the required background on ESN literature, and sets common mathematical grounds for the rest of the work. Section \ref{sec:PropFramework} introduces the framework and analyses each of the techniques proposed in its internal subsections. Section \ref{sec:Experiments} presents the experiments designed to ensure the viability of this study with real data. Section \ref{sec:ResultsDisc} analyzes and discusses the obtained results. Finally Section \ref{sec:ConclusionOutlook} ends the paper by drawing conclusions and outlining future research lines rooted on our findings.

\section{Background} \label{sec:background}

Before proceeding with the description of the proposed suite of XAI techniques, this section briefly revisits the fundamentals of ESN and Deep ESN models (Subsection \ref{ssec:esn}), as well as notable advances in the explainability of recurrent neural networks (Subsection \ref{ssec:RNN}).

\subsection{Echo State Networks: Fundamentals} \label{ssec:esn}

In 2001, Wolfgang Maass and Herbert Jaeger independently introduced Liquid State Machines \cite{maass2002real} and ESNs \cite{jaeger2001echo}, respectively. The combination of these studies with research on computational neuroscience and machine learning \cite{dominey1995complex,steil2004backpropagation} brought up the field of Reservoir Computing. Methods belonging to this field consist of a set of sparsely connected, recurrent neurons capable of mapping high-dimensional sequential data to a low-dimensional space, over which a learning model can be trained to capture patterns that relate this low-dimensional space to a target output. This simple yet effective modeling strategy has been harnessed for regression and classification tasks in a diversity of applications, such as road traffic forecasting \cite{del2020deep}, human recognition \cite{palumbo2016human} or smart grids \cite{crisostomi2015prediction}, among others \cite{gallicchio2017deep}. 

Besides their competitive modeling performance in terms of accuracy/error, Reservoir Computing models are characterized by a less computationally demanding training process than other recursive models: in these systems, only the learner mapping the output of the reservoir to the target variable of interest needs to be trained. Neurons composing the reservoir are initialized at random under some stability constraints. This alternative not only alleviates the computational complexity of recurrent neural networks, but also circumvents one of the downsides of gradient backpropagation, namely, exploding and vanishing gradients. 
\begin{figure}[!ht]
	\vspace{-2mm}
	\centering
	\includegraphics[width=0.7\columnwidth]{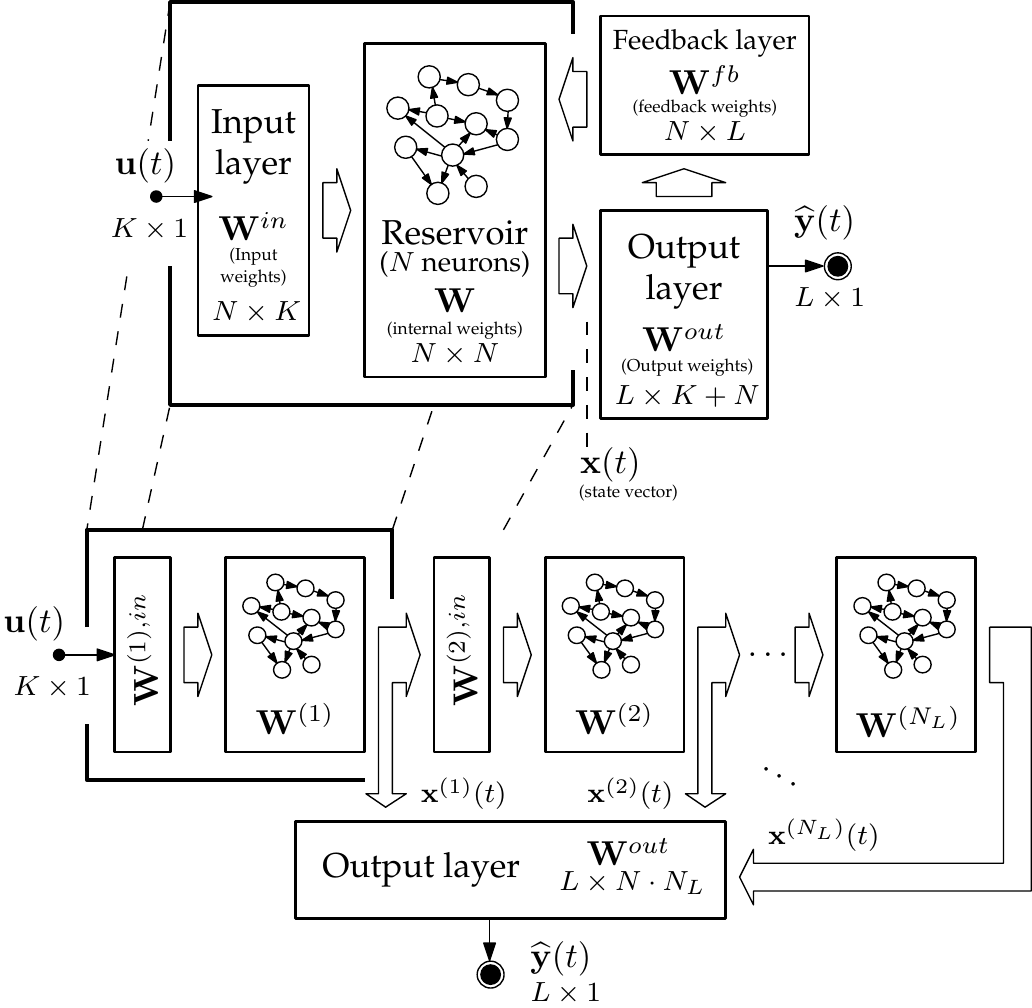}
	\caption{Schematic diagram showing a canonical ESN (upper plot), and the multi-layered stacking architecture of a Deep ESN model (below).}
	\label{fig:ESNConcept}
	\vspace{-1mm}
\end{figure}

To support the subsequent explanation of the proposed XAI techniques, we now define mathematically the internals and procedure followed by ESNs to learn a mapping from a $K$-dimensional input $\mathbf{u}(t)$ (with $t$ denoting index within the sequence) to a $L$-dimensional output $\mathbf{y}(t)$ by a reservoir of $N$ neurons. Following Figure \ref{fig:ESNConcept}, the reservoir state is updated as:
\begin{align} \label{eq:esn_1}
&\mathbf{x}(t+1) = \nonumber \\
&\alpha f(\mathbf{W}^{in} \mathbf{u}(t\hspace{-0.6mm}+\hspace{-0.6mm}1)\hspace{-0.7mm} + \hspace{-0.6mm}\mathbf{W} \mathbf{x}(t)\hspace{-0.6mm} + \hspace{-0.6mm}\mathbf{W}^{fb} \mathbf{y}(t)) + (1\hspace{-0.6mm}-\hspace{-0.6mm}\alpha)\mathbf{x}(t),
\end{align}
where $\mathbf{x}(t)$ denotes the $N$-sized state vector of the reservoir, $f(\cdot)$ denotes an activation function, \smash{$\mathbf{W}_{N\times N}$} is the matrix of internal weights, \smash{$\mathbf{W}^{in}_{N\times K}$} are the input connection weights, and \smash{$\mathbf{W}^{fb}_{N\times L}$} is a feedback connection matrix. Parameter $\alpha\in\mathbb{R}(0,1]$ denotes the \emph{leaking rate}, which allows to set different learning dynamics in the above recurrence \cite{jaeger2007optimization}. The output of the ESN at index $t$ can be computed once the state of the reservoir has been updated, yielding:
\begin{equation} \label{eq:esn_2}
\widehat{\mathbf{y}}(t) = g\left(\mathbf{W}^{out} [\mathbf{x}(t);\mathbf{u}(t)]\right)=g(\mathbf{W}^{out}\mathbf{z}(t)),
\end{equation}
where $[\mathbf{a};\mathbf{b}]$ is a concatenation operator between vectors $\mathbf{a}$ and $\mathbf{b}$, $\mathbf{W}^{out}_{L\times(K+N)}$ is a matrix containing the output weights, and $g(\cdot)$ denotes an activation function. Weights belonging to the aforementioned matrices can be adjusted as per a training data-set with examples $\{(\mathbf{u}(t),\mathbf{y}(t))\}$. However, as opposed to other recurrent neural networks, not all weight inside the matrices $\mathbf{W}^{in}$, $\mathbf{W}$, $\mathbf{W}^{fb}$ and $\mathbf{W}^{out}$ are adjusted. Instead, the weight values of input, hidden state, and feedback matrices are drawn initially at random, whereas those of the output matrix $\mathbf{W}^{out}$ are the only ones tuned during the training phase, using Moore-Penrose pseudo-inversion (for classification) or regularized least squares (regression). In this latter case:
\begin{equation}\label{eq:Wout_reg}
\min\limits_{\mathbf{w}^{out}_{l}} \sum_{t=1}^T \left(\sum_{j=1}^{K+N}\hspace{-0.6mm}w^{out}_{l,j}\cdot z_j(t)\hspace{-0.6mm}-\hspace{-0.6mm}y_l(t)\right)^2\hspace{-2mm}\hspace{-0.6mm}+\hspace{-0.6mm}\lambda \left\|\mathbf{w}^{out}_{l}\right\|_2^2,
\end{equation}
where $l\in\{1,\ldots,L\}$; $\mathbf{w}^{out}_{l}\in\mathbb{R}^{K+N}$; $\left\|\cdot\right\|_2$ denotes $L_2$ norm; $z_j(t)$ is the $j$-th entry of $\mathbf{z}(t)$; and $\mathbf{w}^{out}_{l}=[w^{out}_{l,1},$ $...,w^{out}_{l,K+N}]^T$ denotes the $l$-th row of $\mathbf{W}^{out}$; and $\lambda\in\mathbb{R}[0,\infty)$ permits to adjust the importance of the $L_2$ regularization term in the minimization.

More recently, a multi-layered architecture based on the leaky ESN model described above was introduced in \cite{gallicchio2017deep}. In essence, the Deep ESN model embodies a stacking ensemble of $N_L$ reservoirs, set one after the other, forming a hierarchically layered series of reservoirs. As a result of this concatenation, the state vector of the global Deep ESN model is given by $[\mathbf{x}^{(1)}(t);...;\mathbf{x}^{(N_L)}(t)]$ $\doteq [\mathbf{x}^{(l)}(t)]_{l=1}^{N_L}$, which can be conceived as a multi-scale representation of the input $\mathbf{u}(t)$. It is this property, together with other advantages such as a lower computational burden of their training algorithm and the enriched reservoir dynamics of stacked reservoirs \cite{gallicchio2019richness}, what lends Deep ESNs a competitive modeling performance when compared to other RNNs. 

As shown in Figure \ref{fig:ESNConcept}, in what follows we use $\scriptsize{(l)}$ to indicate that the parameter featuring this index belongs to the $l$-th layer of the Deep ESN. The first stacked ESN is hence updated as:
\begin{align} \label{eq:deepesn_1}
&\mathbf{x}^{(1)}(t+1) = \left(1-\alpha^{(1)}\right)\mathbf{x}^{(1)}(t) + \nonumber \\
&\alpha^{(1)} f\left(\mathbf{W}^{(1)} \mathbf{x}^{(1)}(t) + \mathbf{W}^{(1),in} \mathbf{u}^{(1)}(t+1)\right),
\end{align}
whereas the recurrence in layers $l\in\{2,\ldots,N_L\}$ is given by:
\begin{align} \label{eq:deepesn_2}
&\mathbf{x}^{(l)}(t+1) = \left(1-\alpha^{(l)}\right)\mathbf{x}^{(l)}(t) + \nonumber \\
&\alpha^{(l)} f\left(\mathbf{W}^{(l)} \mathbf{x}^{(l)}(t) + \mathbf{W}^{(l),in} \mathbf{x}^{(l-1)}(t+1)\right),
\end{align}
and the Deep ESN output yields as:
\begin{equation} \label{eq:deepsn_3}
\widehat{\mathbf{y}}(t) = g\left(\mathbf{W}^{out} [\mathbf{x}^{(1)}(t);\ldots;\mathbf{x}^{(N_L)}(t)]\right),
\end{equation}
where $\mathbf{W}_{L\times (N_L \cdot N)}^{out}$ is the weight matrix that maps the concatenation of state vectors of the stacked reservoirs to the target output. Analogously to canonical single-layer ESN, weights in $\mathbf{W}^{(l)}$ and $\mathbf{W}^{(l),in}$ for each layer $l=1,\ldots,N_L$ are initialized at random and rescaled to fulfill the so-called Echo State Property \cite{gallicchio2017echo}, i.e.:
\begin{equation} \label{eq:deepsn_esp}
\max _{l\in\{1,\ldots,N_{L}\}} \rho\left((1-\alpha^{(l)}) \mathbf{I}+\alpha^{(l)} \mathbf{W}^{(l)}\right)<\rho_{max},
\end{equation}
with $\rho(\cdot)$ denoting the largest absolute eigenvalue of the matrix set at its argument, and $\rho_{max}<1$ the so-called \emph{spectral radius} of the model. Once $\mathbf{W}^{(l)}$ and $\mathbf{W}^{(l),in}$ have been set at random fulfilling the above property, they are kept fixed for the rest of the training process, whereas weights in $\mathbf{W}^{out}$ are tuned over the training data by means of regularized least-squares regression as per Expression \eqref{eq:Wout_reg} (or any other low-complexity learning model alike).

Despite the simplicity and computational efficiency of the ESN and Deep ESN training process, the composition of the network itself -- namely, the selection of the number of layers and the value of hyperparameters such as $\alpha^{(l)}$ is a matter of study that remain so far without a clear answer \cite{jaeger2001echo,jaeger2002tutorial,jaeger2005reservoir}. Some automated approaches have been lately proposed, either relying on the study of the frequency spectrum of concatenated reservoirs \cite{gallicchio2018design} or by means of heuristic wrappers \cite{liu2020nonlinear}. However, to the best of our knowledge there is no previous work elaborating on different explainability measures that can be drawn from an already trained Deep ESN to elucidate diverse properties of its captured knowledge. When the audience for which such measures are produced is embodied by machine learning experts, we postulate that the suite of XAI techniques can help them discern what they observe at its input, quantify relevant features (e.g. its memory depth) and thereby, ease the process of configuring them properly for the task at hand.

\subsection{Explainability of Recurrent Neural Networks} \label{ssec:RNN}

Several application domains have traditionally shown a harsh reluctance against embracing the latest advances in machine learning due to the opaque, black-box nature of models emerging over the years. This growing concern with the need for explaining black-box models has been mainly showcased in Deep Learning models for image classification, wherein explanations produced by XAI techniques can be inspected easily. However, the explainability of models developed for sequence data (e.g. time series) has also been studied at a significantly smaller corpus of literature. A notable milestone is the work in \cite{arras2017explaining}, where a post-hoc XAI technique originally developed for image classification was adopted for LSTM architectures, thereby allowing for the generation of local explanations in sequential data modeling and regression. Stimulated in part by this work, several contributions have hitherto proposed different approaches for the explainability of recurrent neural networks, including gradient-based approaches \cite{li2016visualizing,denil2014extraction}, ablation-based estimations of the relevance of particular sequence variables for the predicted output \cite{li2016understanding,kadar2017representation}, or the partial linearization of the activations through the network, permitting to isolate the contribution at different time scales to the output \cite{murdoch2018beyond}. 

Despite the numerous attempts made recently at explaining recurrent neural networks, efforts have been concentrated on explaining networks based on LSTMs and GRU units, thus leaving aside other flavors of recurrent neural computation whose black-box nature also calls for studies on their explainability. Indeed, this noted lack of proposals to elicit explanations for alternative models is in close agreement with the conclusions and prospects made in comprehensive surveys on XAI \cite{arrieta2020explainable,samek2019explainable}. This is the purpose of the set of techniques presented in the following section, to provide different interpretable indicators of the properties of a trained Deep ESN model, as well as information that permit to visualize and summarize the knowledge captured by its layered arrangement of reservoirs.

\section{Proposed Framework} \label{sec:PropFramework}

The framework proposed in this work is composed by three XAI techniques, each tackling some of the most common aspects that arise when training an ESN model or understanding their captured knowledge. These techniques cover three main characteristics that help understand strengths and weaknesses of these models:
\begin{enumerate}[leftmargin=*] 
\item \emph{Potential Memory}, which is a simple test that closely relates to the amount of sequential memory retained within the network. The potential memory can be quantified by addressing how the model behaves when its input fades abruptly. The intuition behind this concept recalls to the notion of \emph{stored information}, i.e., the time needed by the system to return to its resting state.

\item \emph{Temporal Patterns}, which target the visualization of patterns occurring inside the system by means of \emph{recurrence plots} \cite{marwan2007recurrence,eckmann1995recurrence}. Temporal patterns permit to inspect the multiple scales at which patterns are captured by the Deep ESN model through their layers, easing the inspection of the knowledge within the data that is retained by the neuron reservoir. This technique also helps determine when the addition of a layer does not contribute to the predictive performance of the overall model.

\item \emph{Pixel Absence Effect}, which leverages the concept of \emph{local explanations} \cite{arrieta2020explainable} and extends it to ESN and Deep ESN models. This technique computes the prediction difference resulting from the suppression of one of the inputs of the model. The generated matrix of deviations will be contained within the dimensions of the input signal, allowing for the evaluation of the importance of the absence of a single component over the input signal.
\end{enumerate}

In the following subsections we provide rationale for the above set of XAI techniques, their potential uses, and examples with synthetic data previewing the explanatory information they produce.

\subsection{Potential Memory} \label{sec:PropFramework:Tec1}

One of the main concerns when training an ESN model is to evaluate the potential memory the network retains within their reservoir(s). Quantifying this information becomes essential to ensure that the model possesses enough modeling complexity to capture the patterns that best correlate with the target to be predicted. 

Intuitively, a reservoir is able to hold information to interact with the newest inputs being fed to the network in order to generate the current output. It follows that, upon the removal of these new inputs, the network should keep outputting the information retained in the reservoir, until it is flushed out. The multi-layered structure of the network does not allow relating the flow through the Deep ESN model directly to its memory, since the state in which this memory is conserved would most probably reside in an unknown abstract space. This is the purpose of our proposed potential memory indicator: to elicit a hint of the memory contained in the stacked reservoirs. When informed to the audience, the potential memory can improve the confidence of the user with respect to the application at hand, especially in those cases where a hypothesis on the memory that the model should possess can be drawn from intuition. For instance, in an order-ten nonlinear autoregressive-moving average (NARMA) system, the model should be able to store at least ten past steps to produce a forecast for the next time step. 

Before proceeding further, it is important to note that a high potential memory is not sufficient for the model to produce accurate predictions for the task under consideration. However, it is a necessary condition to produce estimations based on inputs further away than the number of sequence steps upper bounded by this indicator. 

For the sake of simplicity, let us assume a single-layer ESN model without leaking rate ($\alpha=1$) nor feedback connection (i.e. $\mathbf{W}^{fb}=\mathbf{0}_{N\times L}$). These assumptions simplify the recurrence in Expression \eqref{eq:esn_1} to:
\begin{align} \label{eq:esn_1_simple}
&\mathbf{x}(t+1) = f(\mathbf{W}^{in} \mathbf{u}(t+1) + \mathbf{W} \mathbf{x}(t)),
\end{align}
which can be seen as an expansion of the current input $\mathbf{u}(t+1)$ and its history represented by $\mathbf{x}(t)$. If we further assume that the network is initially\footnote{For clarity we avoid any consideration to the bias term in this statement.} set to $\widehat{\mathbf{y}}(0)=\mathbf{0}$ when $\mathbf{u}(0)=\mathbf{0}$, the system should evolve gradually to the same state when the input signal $\mathbf{u}(t)$ is set to $\mathbf{0}$ at time $t=T_0$. The time taken for the model to return to its initial state is the potential memory of the network, representing the information of the history kept by the model after training. Mathematically:
\begin{definition}[Potential memory]
Given an already trained ESN model with parameters $$\{\alpha^{(l)},\mathbf{W}^{(l),in},\mathbf{W}^{(l)},\mathbf{W}^{fb},\mathbf{W}^{out}\},$$and initial state set to $\widehat{\mathbf{y}}(0)=\mathbf{y}_0\in\mathbb{R}^L$ when $\mathbf{u}(0)=\mathbf{0}$, the potential memory (PM) of the model at evaluation time $T_0$ is given by:
\begin{equation}
\mbox{PM}({T_0})=T_0-\argmin_{t>T_0} \left\|\widehat{\mathbf{y}}(t)-\mathbf{y}_0\right\|_2 < \epsilon,
\end{equation}
where $\epsilon$ is the tolerance below which convergence of the measure is declared.
\end{definition}
\begin{figure}[!h]
	\vspace{-1mm}
	\centering
	\includegraphics[width=0.8\columnwidth]{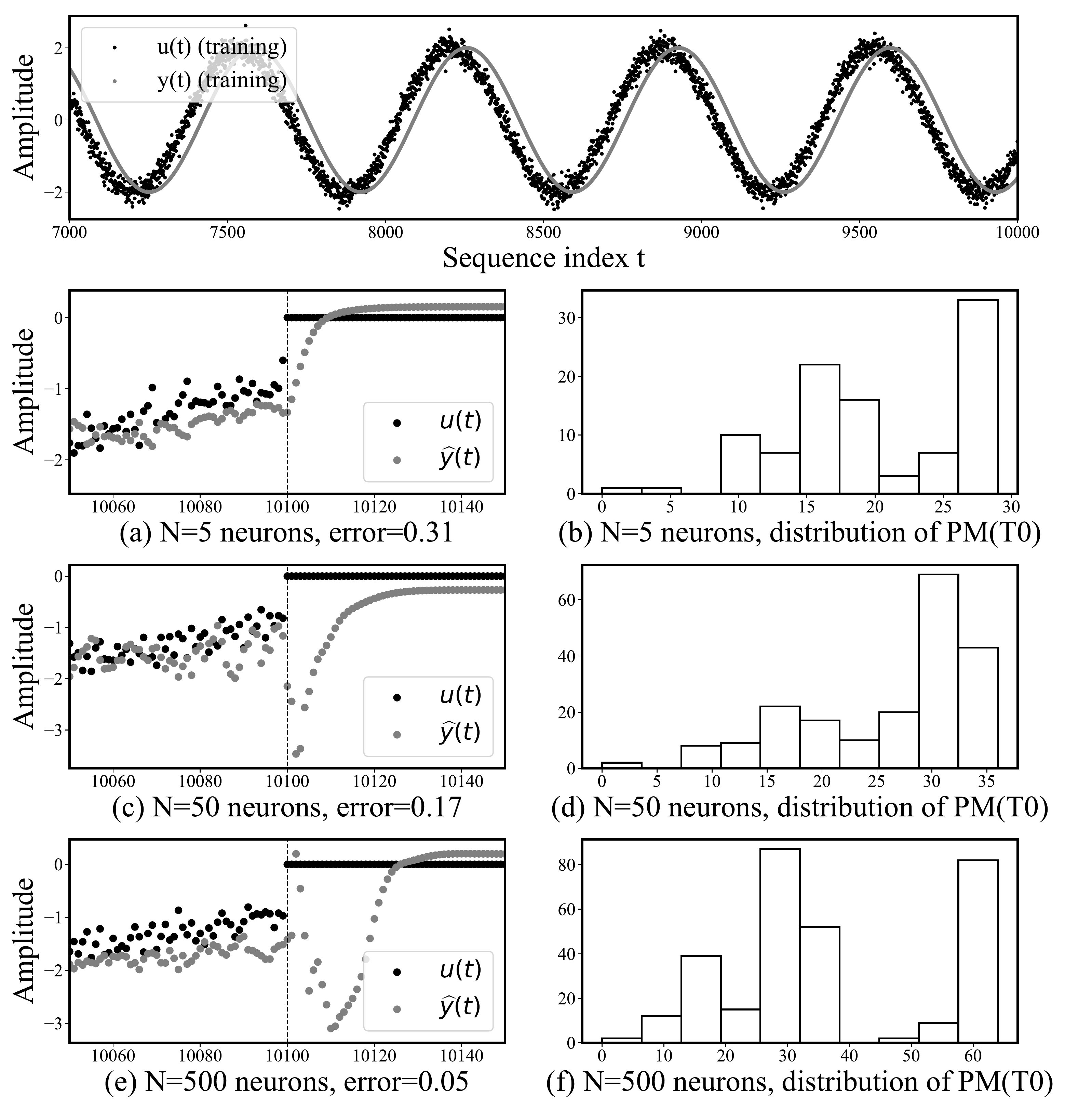}
	\caption{(Top) Input and target sinusoids used for training an ESN model with $\alpha=0.5$ and a single reservoir with $N\in\{5,50,500\}$ neurons. In the rest of plots, reservoir memory dynamics when the input signal is set to $\mathbf{0}$ at $t=10100$, along with the histogram of $\text{PM}(T_0)$ values obtained for $T_0\in (10000,10100]$ and $\epsilon=0.05$.}
	\label{fig:PM}
\end{figure}

In order to illustrate the output of this first technique, we use a single ESN model with varying number of neurons in its reservoir to learn the phase difference of $50$ samples between an input and an output sinusoid. Plots nested in Figure \ref{fig:PM} depict the fade dynamics of different ESN models trained with $N=5$ (plots \ref{fig:PM}.a and \ref{fig:PM}.b), $N=50$ (plots \ref{fig:PM}.c and \ref{fig:PM}.d) and $N=500$ neurons (plots \ref{fig:PM}.e and \ref{fig:PM}.f) in their reservoirs. Specifically, the graph on the top represents the actual signal to be predicted, while the bottom graphs display the behavior of the output of each ESN model when the input signal is zeroed at time $T_0=10100$, along with the empirical distribution of the potential memory computed for $T_0\in (10000,10100]$. 

When the reservoir is composed by just $N=5$ neurons, the potential memory of the network is low given the quick transition of the output signal to its initial state after the input is set to $\mathbf{0}$. When increasing the size of the reservoir to $50$ neurons, a rather different behavior is noted in the output of the ESN model when flushing its contained knowledge, not reaching its steady state until $20$ time steps for the example depicted in Figure \ref{fig:PM}.c. Finally, an ESN comprising 500 neural units in its reservoir does not come with a significant increase of its potential memory, however, the model is able to compute the target output in a more precise fashion with respect to the previous case. This simple experiment shows that the potential memory is a useful metric to evaluate the past range of the sequence modeled and retained at its input. As later exposed in Section \ref{sec:ResultsDisc}, this technique is of great help when fitting a model, since only when the potential memory is large enough, does the model succeed on the testing.

\subsection{Temporal Patterns} \label{sec:PropFramework:Tec2}

One important aspect when designing and building a model is to ascertain whether it has been able to capture the temporal patterns observed in the dynamics of the data being modeled. To shed light on this matter, our devised \emph{temporal patterns} technique resorts to a well-established tool for the analysis of dynamical systems: Recurrence Plots \cite{marwan2007recurrence,eckmann1995recurrence}. This component of the suite aims at tackling two main problems: 1) to determine whether the model has captured any temporal patterns; and 2) if the depth of the model (as per its number of stacked layers) is adding new knowledge that contributes to the predictive performance of the model.

Before proceeding further with the mathematical basis of this technique, we pause at some further motivating intuition behind temporal patterns. Given the black-box nature of neural networks, their training often brings up the question whether the model is capturing the patterns expected to faithfully model the provided data. From logic it can be deduced that, the deeper the network is, the more detail it will be able to hold on to. However, this does not always result in a better predictive model. If we keep in mind that any given layer feeds on the previous layer's high dimension representation (with the exception of the first layer), the patterns found in these latter representations should potentially be more intricate, since their input is already more detailed. 

This is best understood from a simile with the process of examining a footprint. Our human cortex has much definition to extract information from the visualized footprint. However, if we staring at an already enlarged image, by the effect of a magnifying glass, our senses can pick on features they could not detect before, hence obtaining a similar effect to that of layering multiple reservoirs. The depth of the network, as well as the magnification optics, have to be appropriately sized for the task: there is no point in searching for a car using a microscope, as there is no point on using a deep structure to model a simple phenomenon.

Following the findings from \cite{gallicchio2017deep,gallicchio2016deep} in which their authors already found how Deep ESN architectures developed multiple time-scale representations that grew with layering, it seems of utmost importance from an explainability perspective to dispose of a tool that will help us inspect this property in already trained models. For this purpose, we make use of recurrence plots to analyze the temporal patterns found within the different layers of the network. When the layers capture valuable information, their recurrence plots show more focused patterns than their previous layers. This accounts for the fact that each layer is able to better focus on patterns, leaving aside noise present in data. When this growing detail in the recurrence plots disappears at a point of the hierarchy of reservoirs, adding more layers seem counterproductive and does not potentially yield any modeling gain whatsoever.

Mathematically we embrace the definition of recurrence plots in \cite{marwan2007recurrence}. Specifically, the recurrence plot $\mathbf{R}_{t,t'}^{(l,l')}$ between input \smash{$\mathbf{x}^{(l)}(t)$} and \smash{$\mathbf{x}^{(l')}(t')$} is given by:
\begin{equation}
\mathbf{R}_{t,t'}^{(l,l')}=\left\{\begin{array}{lcl}
1 & & \mbox{if }\left\|\mathbf{x}^{(l)}(t)-\mathbf{x}^{(l')}(t')\right\|_2<\epsilon, \\
0 & & \mbox{otherwise,}
\end{array}\right.
\end{equation}
i.e. as a $N\times N$ binary matrix in which we set to value $1$ those indices corresponding to time steps $(t,t')$ where the hidden state vectors of the reservoirs $l$ and $l'$ are equal up to an error $\epsilon$. Obviously, the case when $l=1$ (i.e. the first signal is the input data $\mathbf{u}(t)$), dimensions of the recurrence plot can be enlarged to cover all its length. Likewise, the case with heterogeneously sized reservoirs also fits in the above definition by simply adjusting accordingly the size of the matrix. For our particular case, recurrence plots run for each of the states of a certain layer pair $l,l'$ throughout the test history portrait the repetitions in phase space that happened throughout a given time window. 
\begin{figure}[!ht]
	\vspace{-1mm}
	\centering
	\includegraphics[width=0.67\columnwidth]{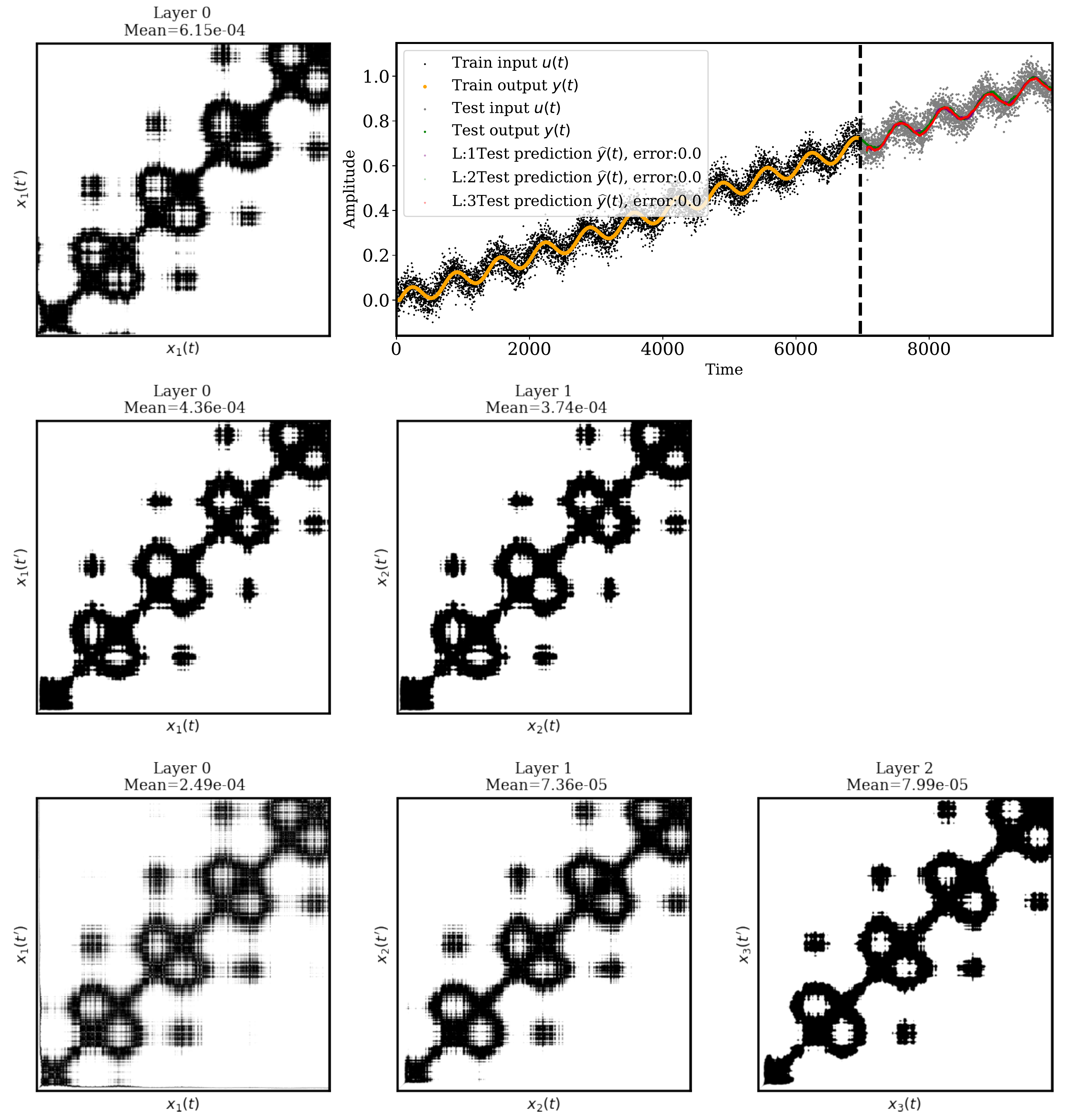}
	\vspace{+1mm}
	\caption{Representation of an ESN model trained to model a trended sinusoid by means of recurrence plots with three different layer configurations (1 layer: top, 2 layers: middle, 3 layers: bottom), displaying the evolution of the patterns seen on the recurrence plots along with an indication of the mean of the recurrence plots.}
	\label{fig:RecurrencePlots}
	\vspace{-1mm}
\end{figure}

Figure \ref{fig:RecurrencePlots} exemplifies the output of the recurrence plots when applied to the output layers of an ESN trained to map a noisy sinusoidal signal with an increasing additive trend to its noiseless version. The figure clearly shows how the recurrent behavior of the signal can be inspected on the recurrence plots of its hidden states. This type of analysis helps detect the properties of the phenomena producing the data under analysis: stationary or non-stationary nature, periodicity, trend, strong fluctuations, similar/inverse evolution through time, static states and similarity between states but at different rates. The extraction of such features from the recurrence plot perfectly couples with the notion of deepening our understanding of these models giving a further insight on the temporal patterns happening within the data. 

This tool serves a two-way purpose that will be further discussed in the experiment section. First, it allows inspecting whether the model is actually capturing the features we could expect from the system; secondly, once these features are captured correctly, it will allow for a better understanding of the system's dynamics through time. For example, the above image clearly shows in each best configuration (two layers) that the model is clearly capturing the system non-stationary condition along with its low amplitude fluctuations that present a recurrent behavior through time. A further analysis of this tool when applied to real data will be covered in Section \ref{sec:ResultsDisc}.
\begin{figure}[!h]
	\vspace{-1mm}
	\centering
	\includegraphics[width=0.7\columnwidth]{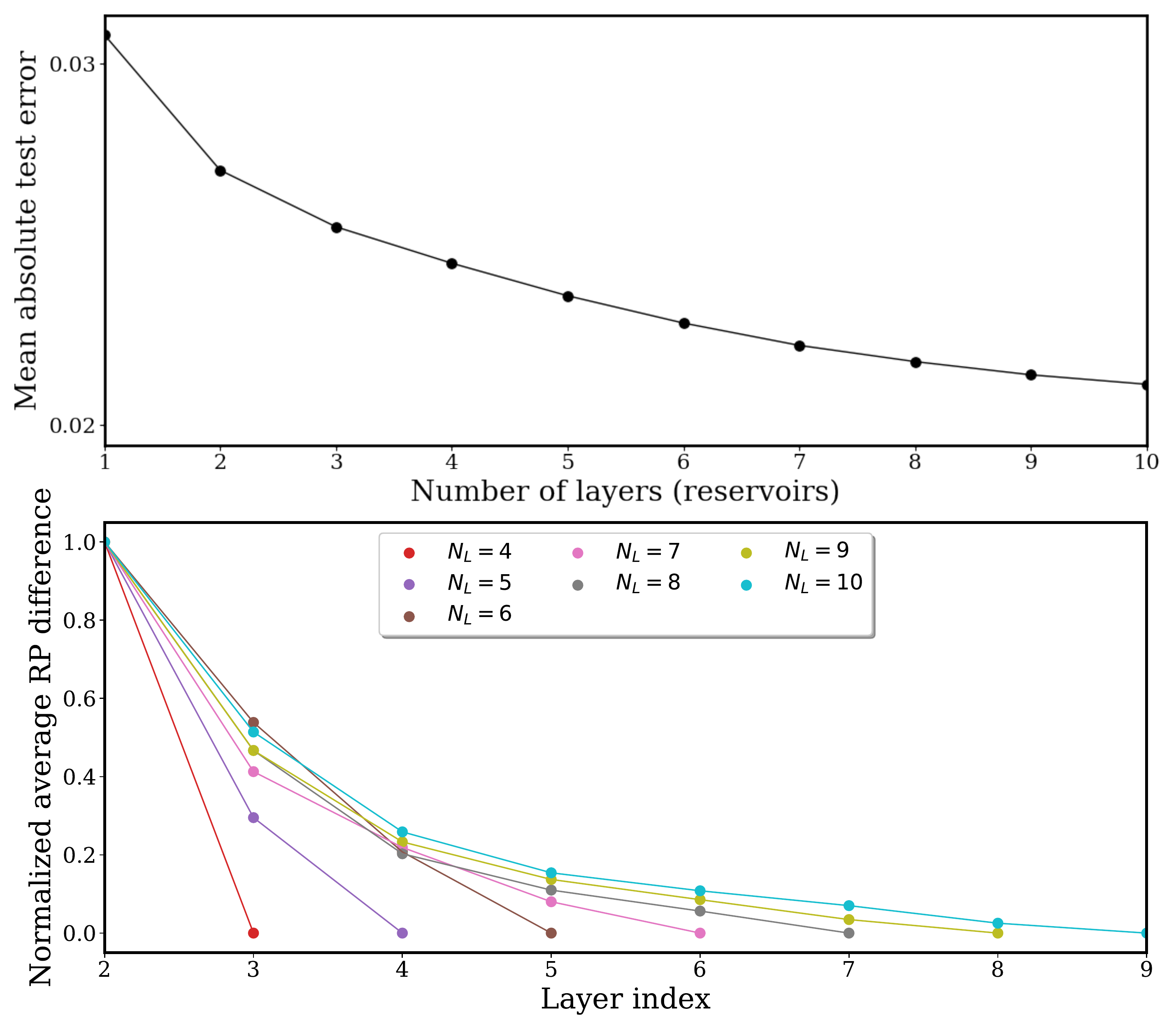}
	\caption{Mean absolute error as a function of the number of reservoirs $N_L$ (top), and evolution of the average difference between recurrence plots corresponding to layers $l-1$ and $l+1$ for $N_L\in\{4,\ldots,10\}$ (bottom). There is a clear positive correlation between the test error and the RP difference, suggesting that performance improvements occur whenever subsequent reservoirs yield a more detailed recurrence plot.}
	\label{fig:LayerUsefulness}
\end{figure}

A byproduct derived this tool is a numerical score that determines whether an added layer is contributing to the modeling process. We devised a preliminary experiment to support the claim that the \emph{sharpness} of a certain layer's recurrent plot adds to the validity of such layer's addition to the predictive power of the model. Given input and output signals, we run the experiment of fitting an ESN of equal characteristics a number of times for each number of layers. For each layer's recurrent plot $R_{t,t'}^{l,l'}$, we calculated its average value, which relates to how sharp the recurrence plot is. Results in Figure \ref{fig:LayerUsefulness} run for the example in Figure \ref{fig:RecurrencePlots} reveals that the average test error computed over $50$ trials correlates with the decrease of the difference between recurrence plots, which validates our intuition about the relative usefulness of every layer.

\subsection{Pixel Absence Effect}\label{sec:PropFramework:Tec3}

Finally, we delve into the third technique that comprises our proposed XAI suite for ESN-based models. We depart from the thought that the analysis of a predictive model always brings up the question of whether an input is causing any effect on the model's prediction. Naturally, we expect to find out how the inputs we deem most important are actually the ones on which the model focuses most when eliciting its predictions. However, discovering whether this is actually true in the trained model is not an easy task.

Intuitively, if we take a step back from the architecture and we simply stare at the difference caused by the cancellation of a certain input, we should be able to observe the individual influence of such input for the issued prediction. Similarly, inputs could be canceled in groups to evaluate the importance of each neighboring region of the input space. Given that the result of this technique can be laid directly over the input of the model, the results can be readily read and understood by any type of user, disregarding his/her technical background. This capability makes this XAI approach promising from an explainability standpoint. Indeed, the term \emph{pixel} in the name of the technique comes purposely from the noted suitability of the technique to be \emph{depicted} along with the input data, applying naturally when dealing with image or video data.

In accordance with the previous notation, we now mathematically describe this technique starting from the input-output a single-layered ESN model in Expression \ref{eq:esn_2}, which we include below for convenience:
\begin{equation}
	\widehat{\mathbf{y}}(t) = g\left(\mathbf{W}^{out} [\mathbf{x}(t);\mathbf{u}(t)]\right)=g(\mathbf{W}^{out}\mathbf{z}(t)).
\end{equation}
The cancellation of a certain point $T_{\boxtimes}$ in the input sequence $\mathbf{u}(t)$ would affect not only in the sequence itself, but also in future predictions $\widehat{\mathbf{y}}(t)$ $\forall t>T_{\boxtimes}$ due to the propagation of the altered value throughout the reservoir. Namely:
\begin{equation}
\widehat{\mathbf{y}}^{\boxtimes}(t)=g\left(\mathbf{W}^{\text {out}}[\mathbf{x}^\boxtimes(t) ; \mathbf{u}^{\boxtimes}(t)]\right),
\end{equation}
where $\mathbf{u}^{\boxtimes}(t)$ denotes the input signal with $\mathbf{u}(T_\boxtimes)=\mathbf{0}_{K\times 1}$, and $\widehat{\mathbf{y}}^\boxtimes(t)$ results from applying recurrence \eqref{eq:esn_1} to this altered signal. The intensity of the effect of the suppression on the output signal at time $T_\boxtimes$ can be quantified as:
\begin{equation}\label{eq:intensity}
\mathbf{e}(t;T_\boxtimes)=[e_l(t;T_\boxtimes)]_{l=1}^L=\widehat{\mathbf{y}}^{\boxtimes}(t)-\widehat{\mathbf{y}}(t),\; \forall t\geq T_{\boxtimes}.
\end{equation}
where $\mathbf{e}(t;T_\boxtimes)\in \mathbb{R}^{L\times 1}$, and its sign indicates the direction in which the output is pushed as per the modification of the input. Clearly, when dealing with classification tasks (i.e. $\mathbf{y}(t)\in\mathbb{N}^{L\times 1}$), a similar rationale can be followed by conceiving the output of the ESN-based model as the class probabilities elicited for the input sequence. Therefore, the intensity computed as per \eqref{eq:intensity} denotes whether the modification of the input sequence \emph{increases} ($e_l(t;T_\boxtimes)>0$) or \emph{decreases} ($e_l(t;T_\boxtimes)<0$) the probability associated to class $l\in\{1,\ldots,L\}$. 

\section{Experimental Setup}\label{sec:Experiments}

We assess the performance of the set of proposed techniques described previously by means of several computer experiments where different ESN-based models are used to model datasets of diverse nature. As such, experiments are divided in three use cases: 
\begin{itemize}[leftmargin=*]
\item We begin with arguably the most explored nature of data in ESN modeling: time series. When dealing with tasks formulated over time series data (particularly forecasting), it is of utmost importance to understand the behavior of the ESN architecture. To this end, we exemplify the output of potential memory and temporal patterns when used to explain an ESN-based model for three different regression tasks: one comprising a variable-amplitude sinusoid, and two real-world datasets for battery consumption and traffic flow forecasting.
 
\item Second, we deal with ESN models used for image classification. This application is not new \cite{woodward2011reservoir,souahlia2016experimental,tong2018reservoir}, but considering this task as a benchmark for our developed XAI tools permits to show their utility to unveil strengths and weaknesses of these architecture in such a setup. 

\item Finally, we undertake video classification with ESN-based models which, to the best of our knowledge, has not been addressed before in the related literature. Differently than the other two tasks under consideration, we will not only analyze the knowledge captured by the ESN-based classifier, but also compare it to a Conv2DLSTM model \cite{shi2015convolutional} used as a baseline for video classification.
\end{itemize}

In the following subsections we describe the setup and datasets used for the tasks described above.

\subsection{Time Series Forecasting}

As mentioned before, time series analysis stands at the core of the ESN architecture's main objective: modeling sequential data. Therefore, three experiments with different time series data sources are devised to gain understanding upon being modeled by means of ESNs. Specifically, we select 1) a synthetic NARMA system of order 10 fed with a variable noisy sinusoidal signal; 2) a dataset composed by real traffic flow data collected by different sensors deployed over the city in Madrid \cite{lana2016role}, and 3) a dataset of lithium-ion battery sensor readings. In all these datasets we formulate a regression task by which the goal is to predict the next value of the time series data, given the history of past values. 
\begin{figure*}[!h]
	\vspace{-1mm}
	\centering
	\includegraphics[width=\textwidth]{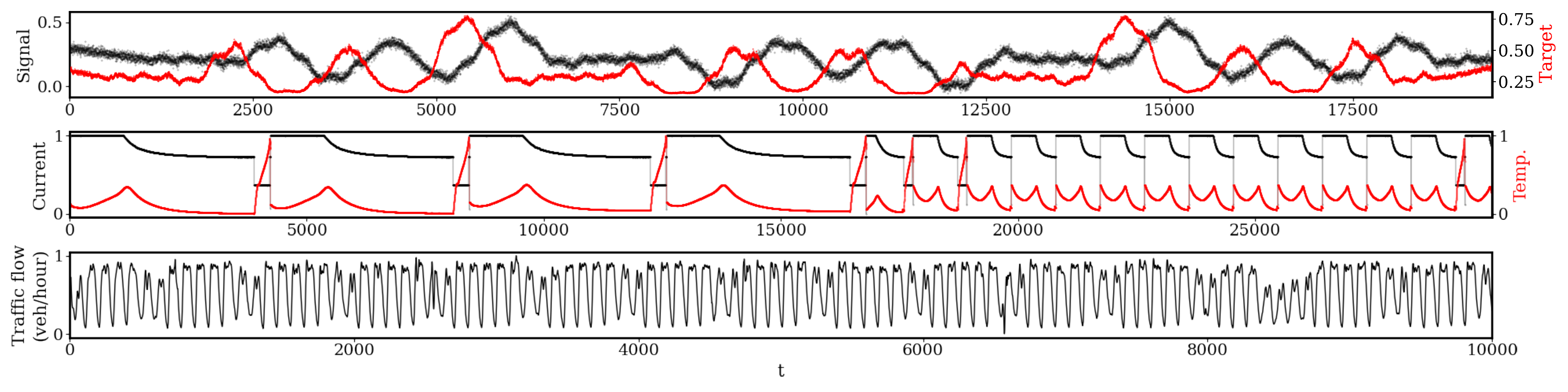}
	\caption{Segments of the different time series datasets used in the experiments: (top) input $u(t)$ (black) and output $y(t)$ (red) of a NARMA model of order 10; (middle) electrical current (black) and temperature (red) from a battery; and (bottom) road traffic flow (vehicles per hour) collected over time.}
	\label{fig:TimeSeries}
	\vspace{-1mm}
\end{figure*}

To begin with, we need a controllable dynamical system producing data to be modeled via a reservoir-based approach. Opting for this initial case ease the process of verifying whether the knowledge of the trained reservoir visualized with our XAI tools conforms to what could be expected from the dynamical properties underneath. For this purpose, we choose a system featuring a recurrent behavior to see its implications on the potential memory and temporal patterns. A NARMA system governed by the following expression meets this recurrent nature sought for the time series:
\begin{equation}\label{eq:narma}
\begin{aligned}
y(t+1)\hspace{-0.75mm}=\hspace{-0.75mm}\tanh(0.3\cdot y(t)+0.05\cdot y(t)\cdot \sum_{i=0}^{9} y(t-i)& \\
+1.5\cdot u(t-9)\cdot u(t)+0.1)&,
\end{aligned}
\end{equation}
i.e., a one-dimensional time series $u(t)$ given by:
\begin{equation}
u(t) = \sin (2\pi F t) +0.02\cdot n(t),
\end{equation}
with $F=4$ Hz, $n(t)\sim \mathcal{N}(0,1)$, i.e. a Gaussian distribution with mean $0$ and unit variance. The recurrence as per Expression \eqref{eq:narma} imposes that any model used for modeling the relationship between $u(t)$ and $y(t+1)$ should focus on at least ten prior time steps when producing its outcome, and should reflect the behavior of the signal amplitude changes over time. A segment of the time series resulting from the NARMA system is shown in Figure \ref{fig:TimeSeries} (top plot).

The second and third datasets, however, deal with the task of forecasting the next value of time series data generated by a real-life recurrent phenomenon. On one hand, a battery dataset is built with recorded electric current and temperature measurements of a lithium-ion battery cycled to depletion following different discharging profiles. This type of data allows for an inspection of a multivariate ESN that presents a very stable working regime, featuring very abrupt changes (as the middle plot in Figure \ref{fig:TimeSeries} clearly shows). On the other hand, traffic flow data collected by a road sensor in Madrid is used. This data source is interesting since it contains long-term recurrence dynamics (bottom plot in Figure \ref{fig:TimeSeries}), suited to be modeled via ESN-based approaches. These datasets have been made available in a public GitHub repository, along with all scripts needed for replicating the experiments (\url{https://github.com/alejandrobarredo/XAI4ESN}).

Methodologically the experiments with the time series datasets discussed in Subsection \ref{ssec:timeseries} are structured in the following way. First, a \emph{potential memory} analysis is carried out for the three datasets, elaborating on how the results relate to the characteristics of the data being modeled. Then, a follow-up analysis is performed around the \emph{temporal patterns} emerging from the reservoir dynamics, also crosschecking whether such patterns conform to intuition. All figures summarizing the outputs of these XAI techniques are depicted together in the same page for a better readability of the results and an easier verification of our claims.
\begin{table*}[ht!]
	\centering
	\caption{List of action recognition video datasets considered in the experiments, along with their characteristics}
	\resizebox{\columnwidth}{!}{%
		\begin{tabular}{lcccL{10cm}}
			\toprule
			Dataset & \# classes & \# videos & Frame size & Description \\
			\midrule
			\texttt{KTH} \cite{schuldt2004recognizing} & 6 & 600 & $160\times120$ & Videos capturing 6 human actions (walking, jogging, running, boxing, hand waving and hand clapping) performed several times by 25 subjects in 4 different scenarios \\
			\midrule
			\texttt{WEIZMANN} \cite{blank2005actions} & 10 & 90 & $180\times 144$ & Videos of 9 different people, each performing 10 natural actions (e.g. run, walk, skip, jumping-jack) divided in examples of 8 frames\\
			\midrule
			\texttt{IXMAS} \cite{weinland2006free} & 13 & 1650 & $320\times 240$ & Lab-generated multi-orientation videos of 5 calibrated and synchronized cameras recording common human actions (e.g. checking-watch, crossing-arms, scratching-head) \\
			\midrule
			\texttt{UCF11} \cite{liu2009recognizing} & 11 & 3040 & \multirow{3}{*}{$320\times 240$} & \multirow{3}{*}{\parbox{10cm}{Action recognition dataset of realistic action videos, collected from YouTube, having a variable number of action categories}} \\
			\texttt{UCF50} \cite{reddy2013recognizing} & 50 & 6676 & & \\
			\texttt{UCF101} \cite{soomro2012ucf101} & 101 & 13320 & & \\
			\midrule
			\texttt{UCFSPORTS} \cite{rodriguez2008action} & 16 & 800 & $320\times 240$ & Sport videos, collected from YouTube, having 16 sport categories \\
			\midrule
			\texttt{HMDB51} \cite{kuehne2011hmdb} & 51 & 6766 & $320\times 240$ & Videos collected from different sources (mainly YouTube and movies) divided into 51 actions that can be grouped in five types: general facial actions, facial actions with object, general body movement, body with object, human body interactions \\
			\bottomrule
		\end{tabular}%
	}
	\label{tab:datasets}
\end{table*}	

\subsection{Image Classification}

As mentioned before, ESN models have been explored for image classification in the past, attaining competitive accuracy levels with a significantly reduced training complexity \cite{woodward2011reservoir,souahlia2016experimental,tong2018reservoir}. To validate whether an ESN-based model is capturing features from its input image that intuitively relate to the target class to be predicted, the second experiment focuses on the application of the developed \emph{pixel absence} technique to an ESN image classifier trained to solve the well-known MNIST digits classification task \cite{lecun1998mnist}. This dataset comprises $60000$ train images and $10000$ test images of $28\times 28$ pixels, each belonging to one among $10$ different categories. 

Since these recurrent models are rather used for modeling sequential data, there is a prior need for encoding image data to sequences, so that the resulting sequence preserves most of the spatial correlation that could be exploited for classification. This being said, we opt for a column encoding strategy, so that the input sequence is built by the columnwise concatenation of the pixels of every image. This process yields, for every image, a one-dimensional sequence of $28\times 28=784$ points, which is input to a Deep ESN model with $N_L=4$ reservoirs of $N=100$ neurons each. Once trained, the Deep ESN model achieves a test accuracy of 86\%, which is certainly lower than CNN models used for the same task, but comes along with a significantly decrease in training complexity. Training and testing over this dataset was performed in less than 5 seconds using a paralelized Python implementation of the Deep ESN model and an i7 2.8 GHz processor with 16GB RAM.

In this case results and the discussion held on them focus on the \emph{pixel absence} effect computed not only for single pixels (namely, points of the converted input sequence), but also to $2\times 2$, $4\times 4$ and $8\times 8$ pixel blobs showing the importance the model granted to regions of the input image with increasing granularity. As elaborated in Section \ref{ssec:imageclass}, the absence effects noted in this dataset exemplify further uses beyond explainability, connecting with the robustness of the model against adversarial attacks.

\subsection{Video Classification}

With this third kind of data we take a further step beyond the state of the art, assessing the performance of ESN-based models for video classification. The transformation of an image to sequences paves the way towards exploring means to follow similar encoding strategies for the classification of stacked images, which essentially gives rise to the structure of a video. 

In its seminal approach, the video classification task comprises not only predicting the class associated to an image (frame), but also aggregating the predictions issued for every frame over the length of every video. These two steps can be performed sequentially or, instead, jointly within the model, using for this purpose assorted algorithmic strategies such as joint learnable frame classification and aggregation models. In our case, similarly to image classification, we encode each video (i.e. a series of frames) to a multidimensional sequence data that embeds the evolution of each pixel of the frames over the length of the video. Specifically, each component $u_k(t)$ of $\mathbf{u}(t)$ is a sequence denoting the value of the $k$-th pixel as a function of the frame index $t$. Therefore, the dimension $K$ of the input $\mathbf{u}(t)$ to the ESN model is equal to the number of pixels of the frames composing the video under analysis. 

In order to assess its predictive accuracy, an ESN-based model with $N_L=4$ reservoirs and $N=500$ neurons each is run over several known video classification datasets, which are thoroughly described in Table \ref{tab:datasets}. However, achieving a fair comparison between other video classification models reported in the literature and our ESN-based model is not straightforward. Most existing counterparts are deep neural networks that exploit assorted video preprocessing strategies (e.g. local space-time features \cite{schuldt2004recognizing} or the so-called universal background model \cite{5459427}) and/or massively pretrained weights \cite{ghadiyaram2019large}, which permit to finely tune the network for the task at hand. Including these ingredients in the benchmark would not allow for a fair attribution of the noted performance gaps to the modeling power of one or another classification model. 

To avoid this problem and focus the scope of the discussion, the ESN-based architecture is compared to a Conv2DLSTM deep neural network \cite{shi2015convolutional} having at its input the original, unprocessed sequence of video frames. The CNN part learns visual features from each of the frames, whereas the LSTM part learns to correlate the learned visual features over time with the target class. We emphasize that models are trained with just the information available in each of the datasets, without data augmentation nor pretraining. This comparison allows comparing the strengths of these raw modeling architectures without any impact of other processing elements along the video classification pipeline. Table \ref{tab:models} summarizes the characteristics of these models under comparison, as well as the parameters of the ESN-based approaches used in the preceding tasks. 	

In regards to explainability, we apply our developed pixel-absence technique to the ESN model trained for video classification to extract insights about the model's learning abilities in this setup. The discussion about the results of the benchmark is presented in Subsection \ref{ssec:videoclass}.
\begin{table}[h!]
	\centering
	\caption{Considered ESN and Conv2DLSTM models.}
	\resizebox{0.6\columnwidth}{!}{%
		\begin{tabular}{lC{1.35cm}C{1.35cm}C{1.35cm}C{1.35cm}}
			\toprule
			Experiment & $N_L$ & $N$ & $\alpha$ & $\rho_{max}$ \\
			\midrule
			NARMA & 1 & 100 & 0.99 & 0.95 \\
			Battery & 5 & 200 & 0.90 & 0.9 \\
			Traffic & 2 & 200 & 0.99 & 0.9 \\
			\midrule
			MNIST & 4 & 100 & 0.95 & 0.9 \\
			\midrule
			Video & 4 & 500 & 0.99 & 0.9\\	
			\midrule		
			Conv2DLSTM & \multicolumn{4}{l}{64 $3\times 3$ kernels + LSTM, tanh activation}\\
			& \multicolumn{4}{l}{+ Dropout (0.2)}\\
			& \multicolumn{4}{l}{+ Dense (256 neurons), ReLu activation}\\
			& \multicolumn{4}{l}{+ Dropout (0.3)}\\
			& \multicolumn{4}{l}{+ Dense(\# classes), softmax output}\\
			& \multicolumn{4}{l}{Optimizer: SGD, $lr=0.001$}\\
			& \multicolumn{4}{l}{Categorical cross-entropy, 20 epochs}\\
			\bottomrule
		\end{tabular}%
	}
	\label{tab:models}
\end{table}	

\section{Results and Discussion}\label{sec:ResultsDisc}

This section goes through and discusses on the results obtained for the experiments introduced previously. The section is structured in three main parts: time series forecasting (Subsection \ref{ssec:timeseries}), image classification (Subsection \ref{ssec:imageclass}), and video classification (Subsection \ref{ssec:videoclass}).

\subsection{Time Series Analysis} \label{ssec:timeseries}

As mentioned before, in this first set of experiments two different studies are carried out. The first one is centered around the potential memory technique, whereas the second one focuses on the temporal patterns method.

\subsubsection{Experiment 1: Potential memory} \label{sec:Exp1}

Reservoir size is one of the most important parameters when designing an ESN-based model, mostly because it is a compulsory albeit not sufficient parameter for the model to be able to model a system's behavior. It is compulsory because there is always a minimum amount of memory required to learn a recurrent behavior, yet it is not sufficient because memory does not model behavior by itself, but only allows for its persistence. In most cases, this value is set manually, at most blindly automated by means of wrappers, without any further interpretable information given of the appropriateness of its finally selected value. This experiment attempts at bringing light into this matter.

For this purpose we experimented with different reservoir sizes when fitting an ESN to the three datasets to be modeled. For every dataset, the reservoir size varies while keeping the rest of the parameters fixed. The rest of the parameters are chosen through experimentation. We choose a spectral radius $\rho_{max}$ equal to $0.95$ since all system has long-term interactions between input and target, and a \textit{leaky rate} of 0.99 to ensure a fast plasticity of the reservoirs.
\begin{figure*}[!ht]
	\vspace{-1mm}
	\centering
	\includegraphics[width=0.9\textwidth]{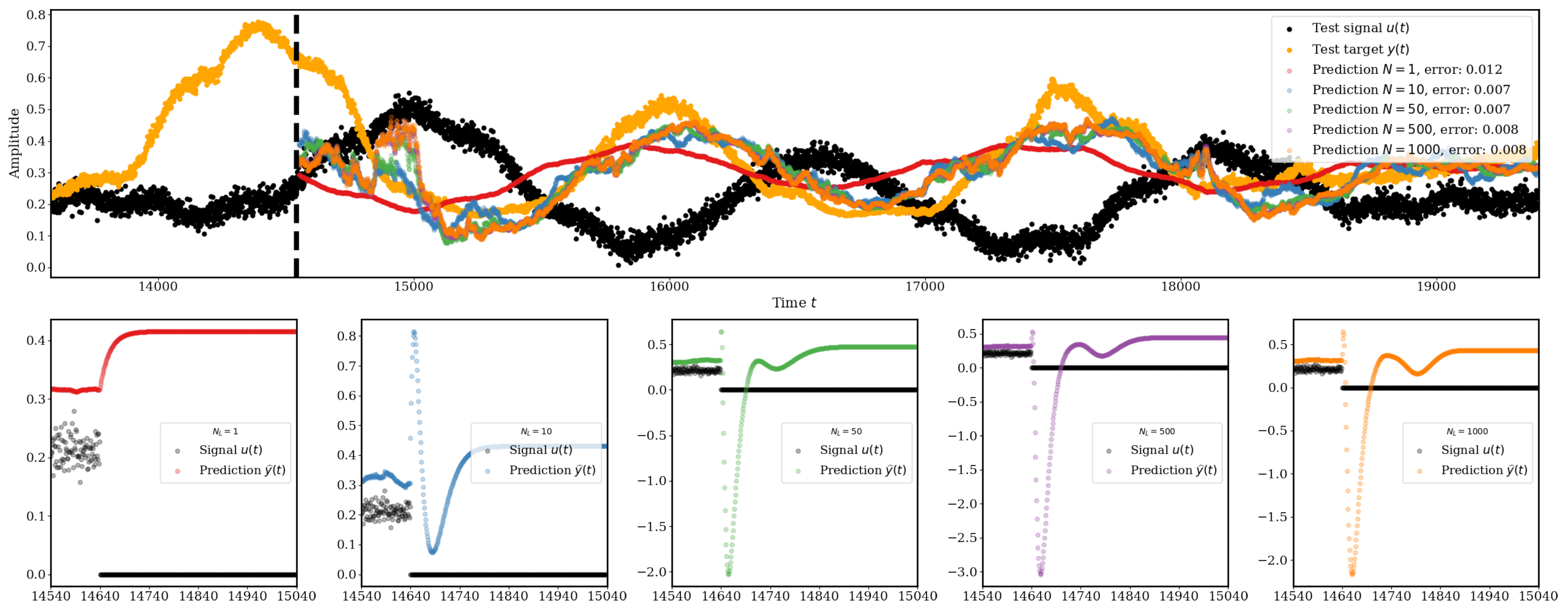}\\
	(a)\\
	\includegraphics[width=0.9\textwidth]{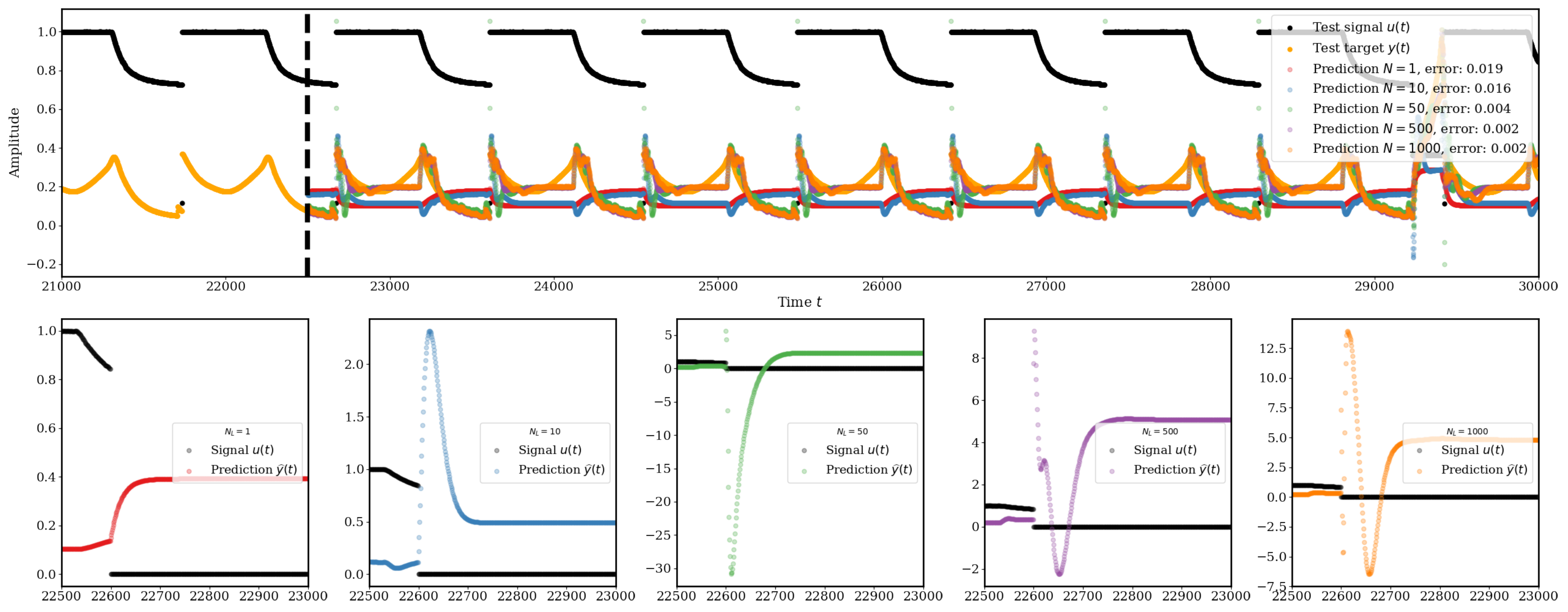}\\
	(b)\\
	\includegraphics[width=0.9\textwidth]{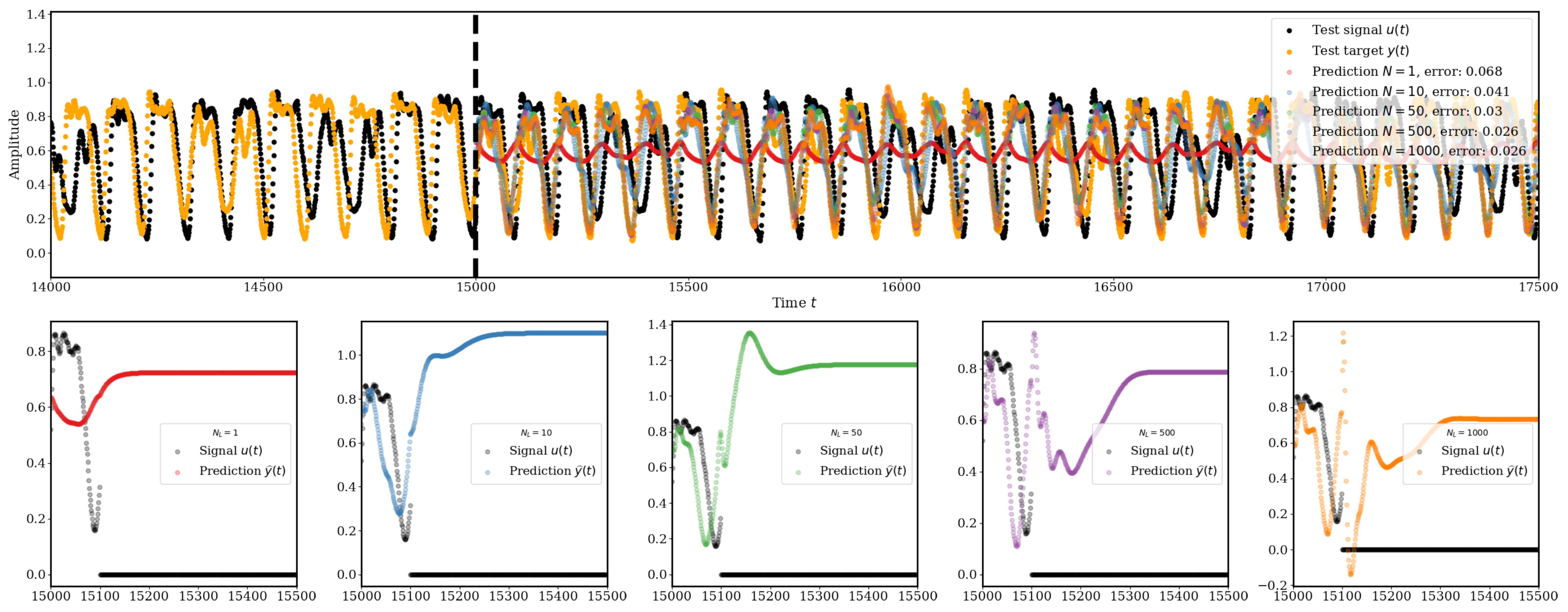}
	\\
	(c)\\
	\caption{Potential memory analysis of the ESN model trained over the (a) NARMA, (b) battery and (c) traffic datasets for different reservoir sizes $N_L=\{1, 10, 50, 500, 1000\}$ and constant reservoir parameters.}
	\label{fig:PM_results}
	\vspace{-1mm}
\end{figure*}

As shown in the images \ref{fig:PM_results}.a, \ref{fig:PM_results}.b and \ref{fig:PM_results}.c, we observe that the size of the reservoir can be considered as a major issue when fitting the model, until a certain threshold is surpassed. The potential memory technique allows monitoring this circumstance by estimating the model's memory capacity. After surpassing a memory threshold, the potential memory of the reservoirs becomes predictable and does not change any longer. This phenomenon seems to be related with its improvement in error rate, although this second aspect should be considered with caution, since the process of fitting an ESN-based model involves several other parameters that interact with each other. In any case, by means of the technique presented in this first study we can observe that there are some hints that might help use understand when we have a workable reservoir size.
\begin{figure*}[!ht]
	\vspace{-1mm}
	\centering
	\includegraphics[width=\textwidth]{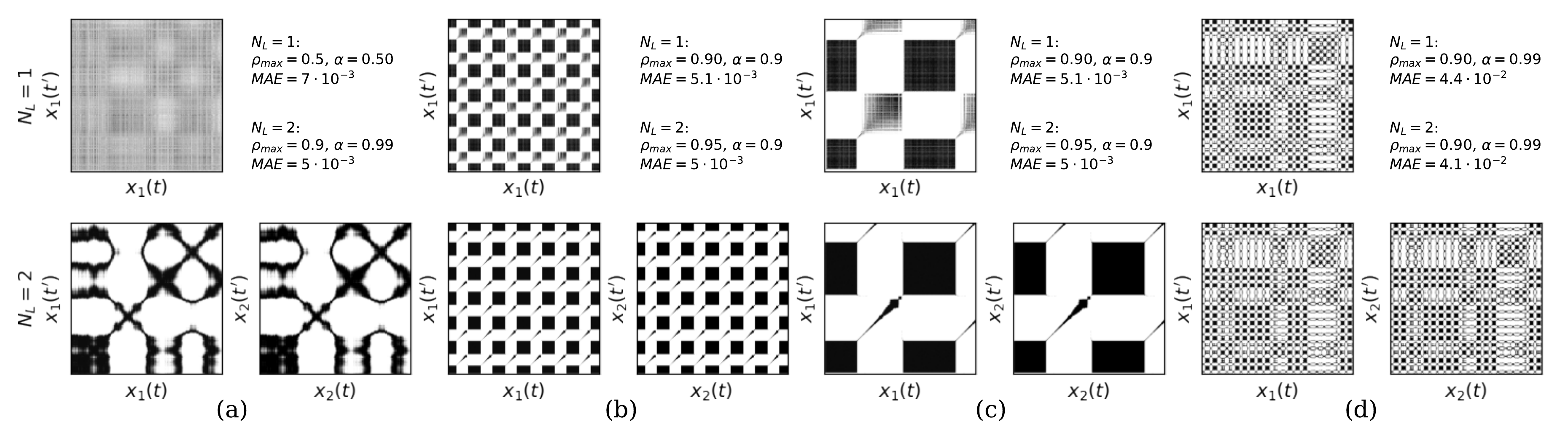}
	\caption{Temporal patterns analysis over the (a) NARMA; (b) and (c) battery; and (d) traffic datasets, presenting different configuration parameters. It can be observed that the spectral radius impact directly on the detail of the recurrence plots, to the extent of causing a significant performance degradation if not selected properly.}
	\label{fig:TP-all}
	\vspace{-1mm}
\end{figure*}

\subsubsection{Experiment 2: Temporal Patterns}\label{sec:Exp2}

When designing a deep ESN model, the next question arising along the process deals with the understanding of whether the model has captured all the dynamics inside the dataset so as to successfully undertake the modeling task. This second experiment analyzes the extent to which a Deep ESN is able to capture the underlying patterns of a system by means of our \textit{temporal patterns}. In this experiment, we vary different parameters to show the usefulness of this technique. This puts in perspective two interesting aspects: first, the technique allows for a quick inspection of whether the model is capturing the patterns of the system. As per the results exposed in what follows, the clarity by which such patterns show up in the plot seems to be related to how well the model is capturing them.

Figures \ref{fig:TP-all}.a, \ref{fig:TP-all}.b/c and \ref{fig:TP-all}.d summarize the results for different reservoir configurations for the three time series datasets under consideration. Two relevant discoveries are found in these plots. In Figure \ref{fig:TP-all}.a (the one eliciting the clearest recurrence plot among the three datasets), we note that the first attempt to model the signal renders very noisy temporal patterns due to a bad selection of the $\rho_{max}$ parameter, although they can still be discerned in the recurrence plot. However, a recurrence plot characterized by more detailed patterns seems to improve its predictive performance, a conclusion that accords with our preliminary insights held around Figure \ref{fig:LayerUsefulness}. In the other two more complex cases (Figures \ref{fig:TP-all}.b/c and \ref{fig:TP-all}.d), we observe that both attempts have successfully modeled the patterns of the system. However, once again the second attempt happens to improve the sensibility of the Deep ESN model to the fine-grained characteristics of the data. This statement is further buttressed in Figure \ref{fig:TP-all}.c, which is an enlargement of a given time window in Figure \ref{fig:TP-all}.b over which a change in the signal is reflected clearly in the recurrence plot.

Apart from the usability of showing interpretable information to the user about the capability of the model to capture the patterns within data, these plots enable another analysis. If we focus on the battery dataset, it is quite clear that the signal is repetitive, stationary and that the system's recurrent patterns are also visible in the recurrence plots. In the case of the traffic data, this phenomenon is much more important. By just staring at the traffic signal it is not easy to realize that the signal also contains certain stationary events (weekends) that produce different behaviors in traffic. By simply staring at the recurrence plots, it is straightforward to realize the periodic structure of the signal, and how these events happen at equally distant time steps throughout the signal (as happens with weekends), with a especially large effect at the top right of the recurrence plot corresponding to a long weekend.

\subsection{Image Classification}\label{ssec:imageclass}

As introduced before, this second experiment analyzes a Deep ESN model for image classification over the MNIST dataset. Once trained, we use this model to examine the output of the third technique comprising our suite of XAI techniques: pixel absence effect.

\begin{figure}[!h]
	\vspace{-1mm}
	\centering
	\includegraphics[width=\columnwidth]{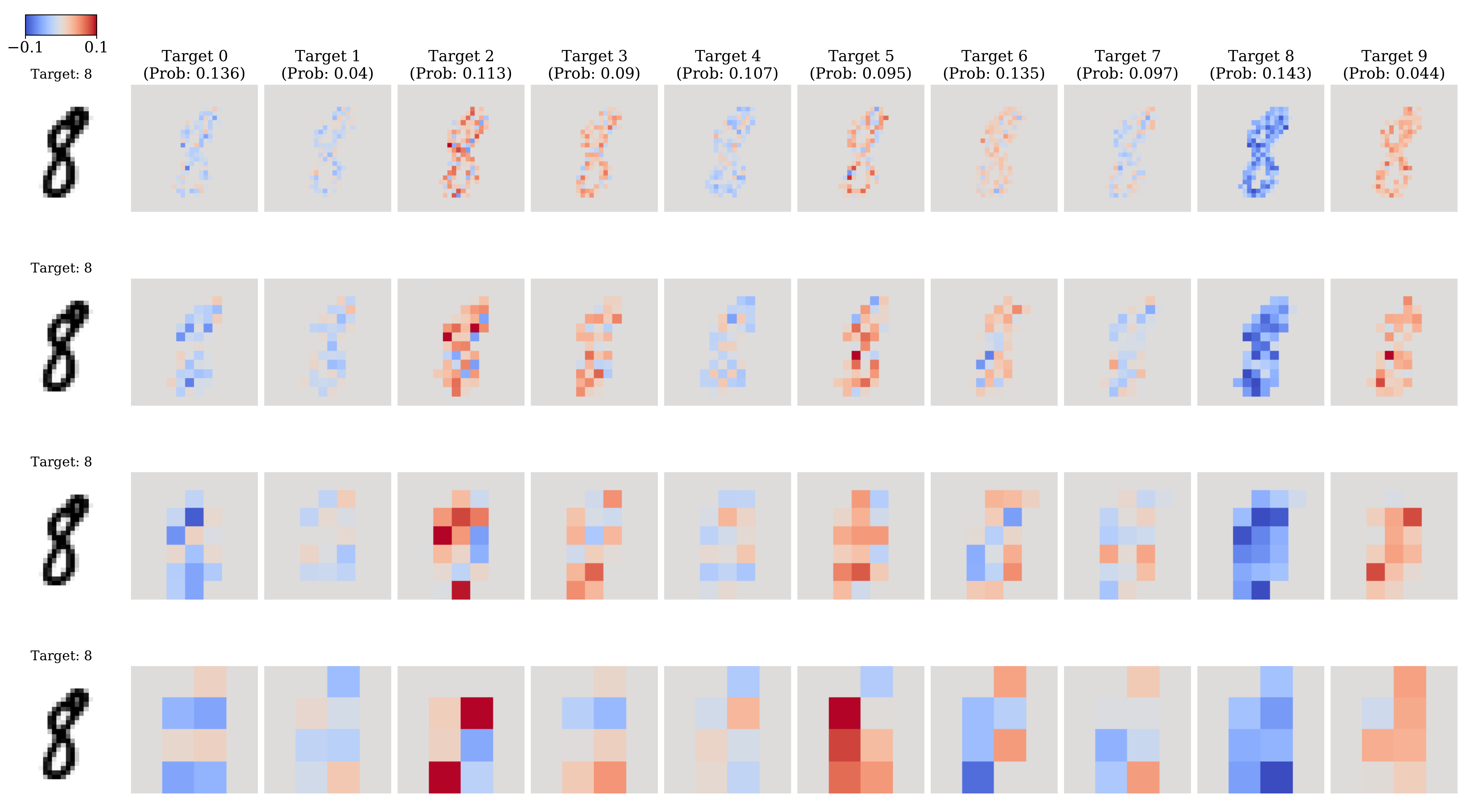}
	\caption{Pixel absence effect over a Deep ESN model trained to classify MNIST digits for single pixels and square $2\times 2$, $4\times 4$ and $8\times 8$ pixel regions.}
	\label{fig:PAE_MNIST}
	\vspace{-1mm}
\end{figure}

Figure \ref{fig:PAE_MNIST} shows the results of computing the pixel absence effect for single pixels, as well as for $2\times 2$, $4\times 4$ and $8\times 8$ pixel blobs. The color represents the importance of the absence of each pixel/blob. Specifically, the scale on the upper left corner shows the amount of change the absence of a certain blob has caused on the probability of every class. The blue color indicates the pixels/blobs that are important to predict the image as the desired class. The darker the blue color of a certain pixel/blob is, the higher the decrease in probability for the desired class will be. The red color means exactly the opposite. Therefore, the output of this technique must be conceived as an adversarially driven local explanation technique, that pursues to explain which regions of a certain input, upon their cancellation, \emph{push} the output class probability distribution towards or against a certain desired class.

We exemplify the output of the experiment for a given example of digit 8. The reason for this choice is that this digit has several coinciding visual features with other digits. As such, we expected to discern a great effect at the center of the digit itself, favoring the classification for digit 0 since the elimination of the center pixels in digit 8 should directly render the image as a 0. However, it is important to note that this does not happen for ESN architectures that are sensitive to space transformations. This sensitivity impedes any ability of the model to extract spatially invariant features from the digit image, thereby circumscribing its knowledge within the sequential columnwise patterns found in examples of its training set. This conclusion emphasizes the relevance of finding transformation strategies from the original domain of data to sequences, not only for ensuring a good performance of the learned ESN-based model, but also to be able to elicit explanations that conform to general intuition.

\subsection{Video Classification}\label{ssec:videoclass}

As stated in the design of the experimental setup, we end up with a video classification task approached via ESN-based models. Since there is no prior work in the literature tackling video classification with ESN architectures, we first run a benchmark to compare its predictive performance to a Conv2DLSTM network, in both cases trained only over unprocessed video data provided in the datasets of Table \ref{tab:datasets}. As argued before, it would not be fair to include preprocessing elements that could bias our conclusions about the modeling power of both approaches under comparison. 
\begin{table}[h!]
	\centering
	\caption{Accuracy and number of trainable parameters of models tested over each video classification dataset.}
	\resizebox{0.85\columnwidth}{!}{%
		\begin{tabular}{L{2cm}ccc}
			\toprule 
			Dataset & Deep ESN & Conv2DLSTM & Difference \\
			\midrule
			\texttt{KTH} & 56\% & 22\% & +34\% \\
			\texttt{WEIZMANN} & 25\% & 12\% & +13\% \\
			\texttt{IXMAS} & 40\% & 36\% & +4\% \\
			\texttt{UCF11} & 54\% & 26\% & +28\% \\
			\texttt{UCF50} & 75\% & 63\% & +12\% \\
			\texttt{UCF101} & 45\% & 42\% & +3\% \\
			\texttt{HMDB51} & 25\% & 17\% & +8\% \\
			\texttt{UCFSPORTS} & 32\% & 23\% & +9\%\\
			\midrule 
			Trainable parameters & \makecell[l]{Min: $12,000$\\Max: $200,000$} & \makecell[l]{Min: $305,626,418$\\Max: $1,205,453,326$} & \makecell{Min: $+293,626,418$\\Max: +$1,005,453,326$}\\
			\bottomrule
		\end{tabular}%
	}
	\label{tab:resultsVideo}
\end{table}

Table \ref{tab:resultsVideo} summarizes the results of our benchmark, both in terms of test predictive accuracy and the size of the trained models (in terms of trainable parameters). It can be observed that Deep ESN models consistently outperforms their Conv2DLSTM counterparts in all the considered datasets, using significantly less trainable parameters that translate to dramatically reduced training latencies. Clearly, these results are far from state-of-the-art levels of performance attained by deep neural networks benefiting from pretrained weights (from e.g. ImageNet). Nevertheless, they serve as an unprecedented preview of the high potential of Deep ESN models to undertake complex tasks with sequence data beyond those for which they have been utilized so far (mostly, time series and text modeling).

Our developed pixel absence effect test was run over the trained Deep ESN model for video classification. We center our discussion on the results for a single video (local explanation) belonging to the \texttt{UCF50} dataset, which should ideally determine what zones of the video frame are most important for the current prediction.
\begin{figure}[!ht]
	\vspace{-1mm}
	\centering
	\includegraphics[width=0.9\columnwidth]{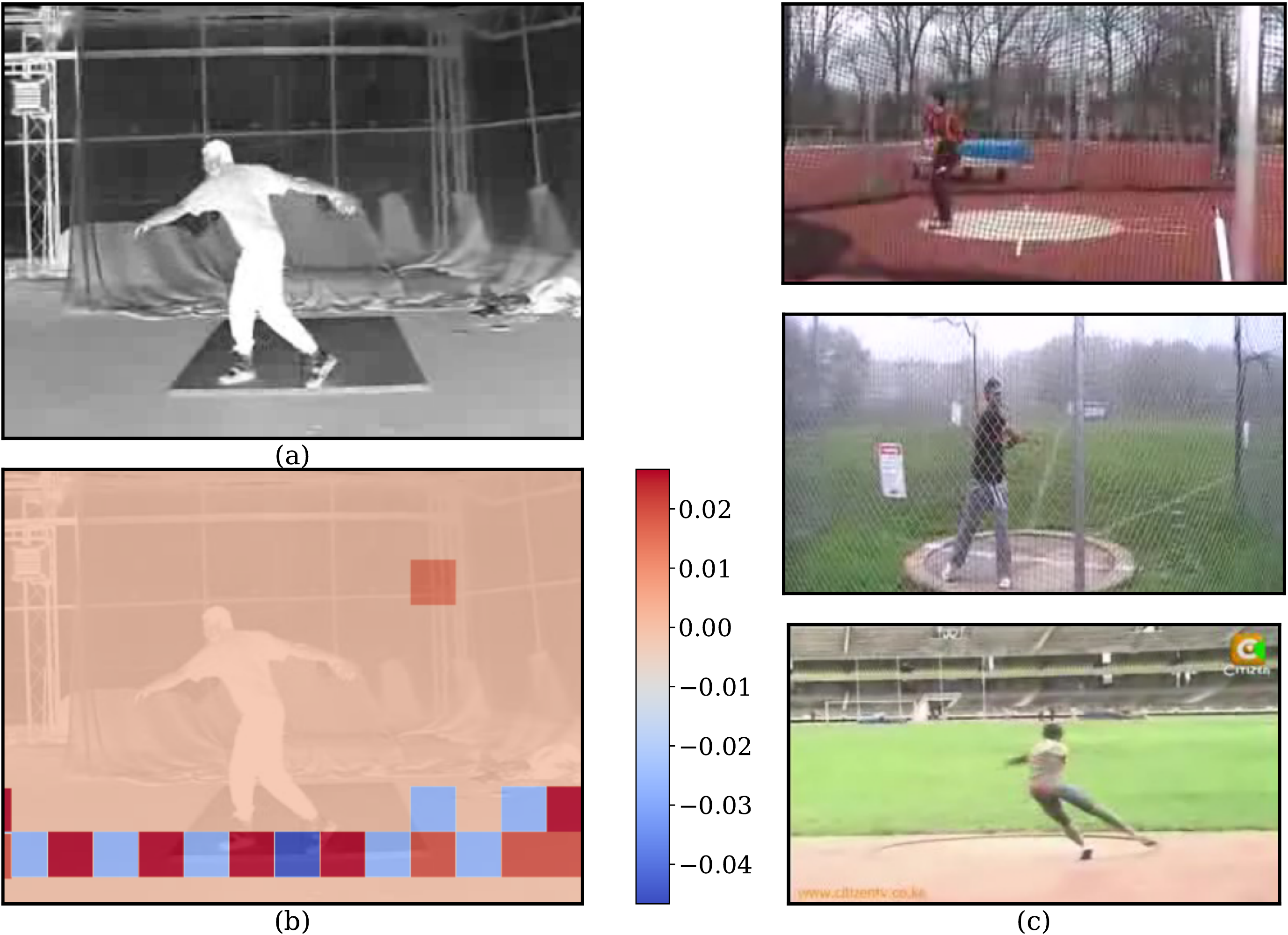}
	\caption{Pixel absence analysis of a video frame showing an unintended result that gives importance to sections of the image that apparently should not be important.}
	\label{fig:PA:Video}
	\vspace{-1mm}
\end{figure}

Interestingly, the results from this pixel absence analysis depicted in Figure \ref{fig:PA:Video} uncovers an unintended yet interesting result. Although the image shown in this picture is just one frame of the video, the region of the frame which is declared to be relevant by the analysis is the same throughout the whole video, the difference being the changing colors of its constituent pixel blobs. This effect unveils that the Deep ESN model has learned to focus on the floor to produce its predictions, which suggests that its knowledge is severely subject to an evident bias in the \texttt{UCF50} dataset. This bias can be detected visually by simply observing the images depicted in Figure \ref{fig:PA:Video}.c, which are samples of frames of the same class that also contain a similar floor feature that might cause a bias in the model. On a closing note, the use of the pixel absent effect tool can help the user detect these sources of bias systematically, so that countermeasures to avoid their effect in the generalization capability of the model can be triggered.

\section{Conclusions and Future Research Lines}\label{sec:ConclusionOutlook}

This manuscript has elaborated on the need for explaining the knowledge modeled over sequential data by ESN-based model by means of a set of novel posthoc XAI techniques. Through the lenses of our proposed methods, we have found out that hidden strengths and weaknesses are present for these type of models. To begin with, the modeling power of these models has been assessed over a diversity of data sources, with results over those that could be expected given their inherent random nature. However, our suite of XAI techniques has also revealed that random reservoirs composed by recurrently connected neural units undergo architectural limitations to model data sources that call for spatially invariant feature learning, such as image and video. Indeed, ESN-based models achieve reasonable levels of predictive accuracy given their low training complexity, specially for video classification without any prior preprocessing, data augmentation and/or pretraining stages along the pipeline. A deeper inspection of their learned knowledge by means of our XAI tools has identified huge data biases across classes in video data, showing that such superior scores might not be extrapolable to videos recorded in other contextual settings. 

All in all, our conclusions can be summarized in the following three statements:
\begin{enumerate}[leftmargin=*]

\item Our tools allow visualizing the amount of temporal memory, the patterns captured by the reservoirs and the most relevant regions of the input to the model in an understandable manner for the general audience. 

\item As shown for image classification, the need for modeling data correlations over space imposes a large dependence of reservoir computing with respect to the strategy used for encoding spatial data as sequences. 

\item In line with the prescriptions and recommendations elevated under the current momentum of the XAI paradigm, we advocate for a major inclusion of information about explainability in modeling exercises with ESN-based models. The adoption of XAI frameworks like the one proposed in this work can be of great interest when auditing models claimed to perform accurately, yet subject to hidden data biases that lead to misleading conclusions about their generalization. 
\end{enumerate}

Several research paths are planned for the future departing from our findings. To begin with, we will investigate different ways to leverage the design flexibility of the reservoirs in a Deep ESN model towards inducing expert knowledge contained in a transparent model in its initialized parameters, so that the propagation of data bias through the model is lessened. In this regard, we foresee to study elements from model distillation to convey such knowledge, possibly by driving the reservoir initialization process not entirely at random. Another interesting research direction is to derive new strategies to transform spatially correlated data into sequences. To this end, we will explore whether neural processing elements in the reservoir can be implemented as convolutional operations over their input data, so that neurons in the reservoir connect different spatial features over time. Finally, a close look will be taken at the interplay between explainability and epistemic uncertainty, the latter especially present in reservoir computing models. As in other randomization based machine learning approaches for sequential data, it is a matter of describing memoirs from the past in a statistically consistent, understandable fashion.

\section*{Acknowledgements}

This work has received funding support from the Basque Government (\emph{Eusko Jaurlaritza}) through the Consolidated Research Group MATHMODE (IT1294-19), EMAITEK and ELKARTEK programs (3KIA project, KK-2020/00049). 

\bibliography{references}

\begin{thebibliography}{10}

\bibitem{jaeger2003adaptive}
Herbert Jaeger.
\newblock Adaptive nonlinear system identification with echo state networks.
\newblock In {\em Advances in neural information processing systems}, pages
  609--616, 2003.

\bibitem{lukovsevivcius2009reservoir}
Mantas Luko{\v{s}}evi{\v{c}}ius and Herbert Jaeger.
\newblock Reservoir computing approaches to recurrent neural network training.
\newblock {\em Computer Science Review}, 3(3):127--149, 2009.

\bibitem{jaeger2004harnessing}
Herbert Jaeger and Harald Haas.
\newblock Harnessing nonlinearity: Predicting chaotic systems and saving energy
  in wireless communication.
\newblock {\em science}, 304(5667):78--80, 2004.

\bibitem{wu2018statistical}
Qiuyi Wu, Ernest Fokoue, and Dhireesha Kudithipudi.
\newblock On the statistical challenges of echo state networks and some
  potential remedies.
\newblock {\em arXiv preprint arXiv:1802.07369}, 2018.

\bibitem{jaeger2005reservoir}
Herbert Jaeger.
\newblock Reservoir riddles: Suggestions for echo state network research.
\newblock In {\em Proceedings. 2005 IEEE International Joint Conference on
  Neural Networks, 2005.}, volume~3, pages 1460--1462. IEEE, 2005.

\bibitem{thiede2019gradient}
Luca~Anthony Thiede and Ulrich Parlitz.
\newblock Gradient based hyperparameter optimization in echo state networks.
\newblock {\em Neural Networks}, 115:23--29, 2019.

\bibitem{ozturk2020optimizing}
Muhammed~Maruf {\"O}zt{\"u}rk, {\.I}brahim~Arda Cankaya, and Deniz {\.I}pekci.
\newblock Optimizing echo state network through a novel fisher maximization
  based stochastic gradient descent.
\newblock {\em Neurocomputing}, 2020.

\bibitem{arrieta2020explainable}
Alejandro~Barredo Arrieta, Natalia D{\'\i}az-Rodr{\'\i}guez, Javier Del~Ser,
  Adrien Bennetot, Siham Tabik, Alberto Barbado, Salvador Garc{\'\i}a, Sergio
  Gil-L{\'o}pez, Daniel Molina, Richard Benjamins, et~al.
\newblock Explainable artificial intelligence ({XAI}): Concepts, taxonomies,
  opportunities and challenges toward responsible ai.
\newblock {\em Information Fusion}, 58:82--115, 2020.

\bibitem{gallicchio2017deep}
Claudio Gallicchio and Alessio Micheli.
\newblock Deep echo state network (deepesn): A brief survey.
\newblock {\em arXiv preprint arXiv:1712.04323}, 2017.

\bibitem{maass2002real}
Wolfgang Maass, Thomas Natschl{\"a}ger, and Henry Markram.
\newblock Real-time computing without stable states: A new framework for neural
  computation based on perturbations.
\newblock {\em Neural computation}, 14(11):2531--2560, 2002.

\bibitem{jaeger2001echo}
Herbert Jaeger.
\newblock The “echo state” approach to analysing and training recurrent
  neural networks-with an erratum note.
\newblock {\em Bonn, Germany: German National Research Center for Information
  Technology GMD Technical Report}, 148(34):13, 2001.

\bibitem{dominey1995complex}
Peter~F Dominey.
\newblock Complex sensory-motor sequence learning based on recurrent state
  representation and reinforcement learning.
\newblock {\em Biological cybernetics}, 73(3):265--274, 1995.

\bibitem{steil2004backpropagation}
Jochen~J Steil.
\newblock Backpropagation-decorrelation: online recurrent learning with o (n)
  complexity.
\newblock In {\em 2004 IEEE International Joint Conference on Neural Networks
  (IEEE Cat. No. 04CH37541)}, volume~2, pages 843--848. IEEE, 2004.

\bibitem{del2020deep}
Javier Del~Ser, Ibai Lana, Eric~L Manibardo, Izaskun Oregi, Eneko Osaba,
  Jesus~L Lobo, Miren~Nekane Bilbao, and Eleni~I Vlahogianni.
\newblock Deep echo state networks for short-term traffic forecasting:
  Performance comparison and statistical assessment.
\newblock In {\em IEEE International Conference on Intelligent Transportation
  Systems (ITSC)}, pages 1--6. IEEE, 2020.

\bibitem{palumbo2016human}
Filippo Palumbo, Claudio Gallicchio, Rita Pucci, and Alessio Micheli.
\newblock Human activity recognition using multisensor data fusion based on
  reservoir computing.
\newblock {\em Journal of Ambient Intelligence and Smart Environments},
  8(2):87--107, 2016.

\bibitem{crisostomi2015prediction}
Emanuele Crisostomi, Claudio Gallicchio, Alessio Micheli, Marco Raugi, and
  Mauro Tucci.
\newblock Prediction of the italian electricity price for smart grid
  applications.
\newblock {\em Neurocomputing}, 170:286--295, 2015.

\bibitem{jaeger2007optimization}
Herbert Jaeger, Mantas Luko{\v{s}}evi{\v{c}}ius, Dan Popovici, and Udo Siewert.
\newblock Optimization and applications of echo state networks with
  leaky-integrator neurons.
\newblock {\em Neural networks}, 20(3):335--352, 2007.

\bibitem{gallicchio2019richness}
Claudio Gallicchio and Alessio Micheli.
\newblock Richness of deep echo state network dynamics.
\newblock In {\em International Work-Conference on Artificial Neural Networks},
  pages 480--491, 2019.

\bibitem{gallicchio2017echo}
Claudio Gallicchio and Alessio Micheli.
\newblock Echo state property of deep reservoir computing networks.
\newblock {\em Cognitive Computation}, 9(3):337--350, 2017.

\bibitem{jaeger2002tutorial}
Herbert Jaeger.
\newblock {\em Tutorial on training recurrent neural networks, covering BPPT,
  RTRL, EKF and the" echo state network" approach}, volume~5.
\newblock GMD-Forschungszentrum Informationstechnik Bonn, 2002.

\bibitem{gallicchio2018design}
Claudio Gallicchio, Alessio Micheli, and Luca Pedrelli.
\newblock Design of deep echo state networks.
\newblock {\em Neural Networks}, 108:33--47, 2018.

\bibitem{liu2020nonlinear}
Kai Liu and Jie Zhang.
\newblock Nonlinear process modelling using echo state networks optimised by
  covariance matrix adaption evolutionary strategy.
\newblock {\em Computers \& Chemical Engineering}, 135:106730, 2020.

\bibitem{arras2017explaining}
Leila Arras, Gr{\'e}goire Montavon, Klaus-Robert M{\"u}ller, and Wojciech
  Samek.
\newblock Explaining recurrent neural network predictions in sentiment
  analysis.
\newblock In {\em Proceedings of the 8th Workshop on Computational Approaches
  to Subjectivity, Sentiment and Social Media Analysis}, pages 159--168, 2017.

\bibitem{li2016visualizing}
Jiwei Li, Xinlei Chen, Eduard Hovy, and Dan Jurafsky.
\newblock Visualizing and understanding neural models in nlp.
\newblock In {\em Proceedings of NAACL-HLT}, pages 681--691, 2016.

\bibitem{denil2014extraction}
Misha Denil, Alban Demiraj, and Nando De~Freitas.
\newblock Extraction of salient sentences from labelled documents.
\newblock {\em arXiv preprint arXiv:1412.6815}, 2014.

\bibitem{li2016understanding}
Jiwei Li, Will Monroe, and Dan Jurafsky.
\newblock Understanding neural networks through representation erasure.
\newblock {\em arXiv preprint arXiv:1612.08220}, 2016.

\bibitem{kadar2017representation}
Akos K{\'a}d{\'a}r, Grzegorz Chrupa{\l}a, and Afra Alishahi.
\newblock Representation of linguistic form and function in recurrent neural
  networks.
\newblock {\em Computational Linguistics}, 43(4):761--780, 2017.

\bibitem{murdoch2018beyond}
W~James Murdoch, Peter~J Liu, and Bin Yu.
\newblock Beyond word importance: Contextual decomposition to extract
  interactions from lstms.
\newblock {\em arXiv preprint arXiv:1801.05453}, 2018.

\bibitem{samek2019explainable}
Wojciech Samek, Gr{\'e}goire Montavon, Andrea Vedaldi, Lars~Kai Hansen, and
  Klaus-Robert M{\"u}ller.
\newblock {\em Explainable AI: interpreting, explaining and visualizing deep
  learning}, volume 11700.
\newblock Springer Nature, 2019.

\bibitem{marwan2007recurrence}
Norbert Marwan, M~Carmen Romano, Marco Thiel, and J{\"u}rgen Kurths.
\newblock Recurrence plots for the analysis of complex systems.
\newblock {\em Physics reports}, 438(5-6):237--329, 2007.

\bibitem{eckmann1995recurrence}
JP~Eckmann, S~Oliffson Kamphorst, D~Ruelle, et~al.
\newblock Recurrence plots of dynamical systems.
\newblock {\em World Scientific Series on Nonlinear Science Series A},
  16:441--446, 1995.

\bibitem{gallicchio2016deep}
Claudio Gallicchio and Alessio Micheli.
\newblock Deep reservoir computing: A critical analysis.
\newblock In {\em ESANN}, 2016.

\bibitem{woodward2011reservoir}
Alexander Woodward and Takashi Ikegami.
\newblock A reservoir computing approach to image classification using coupled
  echo state and back-propagation neural networks.
\newblock In {\em International conference image and vision computing,
  Auckland, New Zealand}, pages 543--458, 2011.

\bibitem{souahlia2016experimental}
Abdelkerim Souahlia, Ammar Belatreche, Abdelkader Benyettou, and Kevin Curran.
\newblock An experimental evaluation of echo state network for colour image
  segmentation.
\newblock In {\em 2016 International Joint Conference on Neural Networks
  (IJCNN)}, pages 1143--1150. IEEE, 2016.

\bibitem{tong2018reservoir}
Zhiqiang Tong and Gouhei Tanaka.
\newblock Reservoir computing with untrained convolutional neural networks for
  image recognition.
\newblock In {\em 2018 24th International Conference on Pattern Recognition
  (ICPR)}, pages 1289--1294. IEEE, 2018.

\bibitem{shi2015convolutional}
Xingjian Shi, Zhourong Chen, Hao Wang, Dit-Yan Yeung, Wai-Kin Wong, and
  Wang-chun Woo.
\newblock Convolutional lstm network: A machine learning approach for
  precipitation nowcasting.
\newblock {\em arXiv preprint arXiv:1506.04214}, 2015.

\bibitem{lana2016role}
Ibai La{\~n}a, Javier Del~Ser, Ales Padr{\'o}, Manuel V{\'e}lez, and Carlos
  Casanova-Mateo.
\newblock The role of local urban traffic and meteorological conditions in air
  pollution: A data-based case study in madrid, spain.
\newblock {\em Atmospheric Environment}, 145:424--438, 2016.

\bibitem{schuldt2004recognizing}
Christian Schuldt, Ivan Laptev, and Barbara Caputo.
\newblock Recognizing human actions: a local svm approach.
\newblock In {\em Proceedings of the 17th International Conference on Pattern
  Recognition, 2004. ICPR 2004.}, volume~3, pages 32--36. IEEE, 2004.

\bibitem{blank2005actions}
Moshe Blank, Lena Gorelick, Eli Shechtman, Michal Irani, and Ronen Basri.
\newblock Actions as space-time shapes.
\newblock In {\em Tenth IEEE International Conference on Computer Vision
  (ICCV'05) Volume 1}, volume~2, pages 1395--1402. IEEE, 2005.

\bibitem{weinland2006free}
Daniel Weinland, Remi Ronfard, and Edmond Boyer.
\newblock Free viewpoint action recognition using motion history volumes.
\newblock {\em Computer vision and image understanding}, 104(2-3):249--257,
  2006.

\bibitem{liu2009recognizing}
Jingen Liu, Jiebo Luo, and Mubarak Shah.
\newblock Recognizing realistic actions from videos ``in the wild''.
\newblock In {\em 2009 IEEE Conference on Computer Vision and Pattern
  Recognition}, pages 1996--2003. IEEE, 2009.

\bibitem{reddy2013recognizing}
Kishore~K Reddy and Mubarak Shah.
\newblock Recognizing 50 human action categories of web videos.
\newblock {\em Machine vision and applications}, 24(5):971--981, 2013.

\bibitem{soomro2012ucf101}
Khurram Soomro, Amir~Roshan Zamir, and Mubarak Shah.
\newblock {UCF101}: A dataset of 101 human actions classes from videos in the
  wild.
\newblock {\em arXiv preprint arXiv:1212.0402}, 2012.

\bibitem{rodriguez2008action}
Mikel~D Rodriguez, Javed Ahmed, and Mubarak Shah.
\newblock Action mach a spatio-temporal maximum average correlation height
  filter for action recognition.
\newblock In {\em 2008 IEEE conference on computer vision and pattern
  recognition}, pages 1--8. IEEE, 2008.

\bibitem{kuehne2011hmdb}
Hildegard Kuehne, Hueihan Jhuang, Est{\'\i}baliz Garrote, Tomaso Poggio, and
  Thomas Serre.
\newblock Hmdb: a large video database for human motion recognition.
\newblock In {\em 2011 International conference on computer vision}, pages
  2556--2563. IEEE, 2011.

\bibitem{lecun1998mnist}
Yann LeCun.
\newblock The mnist database of handwritten digits.
\newblock {\em http://yann.lecun.com/exdb/mnist/}, 1998.

\bibitem{5459427}
{Dong Han}, {Liefeng Bo}, and C.~{Sminchisescu}.
\newblock Selection and context for action recognition.
\newblock In {\em 2009 IEEE 12th International Conference on Computer Vision},
  pages 1933--1940, 2009.

\bibitem{ghadiyaram2019large}
Deepti Ghadiyaram, Du~Tran, and Dhruv Mahajan.
\newblock Large-scale weakly-supervised pre-training for video action
  recognition.
\newblock In {\em Proceedings of the IEEE/CVF Conference on Computer Vision and
  Pattern Recognition}, pages 12046--12055, 2019.

\end{thebibliography}

\end{document}